\documentclass[10pt,twocolumn,letterpaper]{article}

\usepackage[dvipsnames]{xcolor}

\usepackage{cvpr}              
\newcommand{\tcr}[1]{\textcolor{black}{#1}}
\usepackage{graphicx}
\usepackage{amsmath}
\usepackage{amssymb}
\usepackage{tabularx,booktabs}
\usepackage{makecell}
\usepackage{xr}
\usepackage{float}
\newcolumntype{C}{>{\centering\arraybackslash}X} 
\usepackage{lipsum}
\usepackage[none]{hyphenat}
\usepackage{multirow}
\usepackage{csquotes}
\usepackage[accsupp]{axessibility}

\definecolor{cvprblue}{rgb}{0.21,0.49,0.74}
\usepackage[pagebackref,breaklinks,colorlinks]{hyperref}

\usepackage[capitalize]{cleveref}
\crefname{section}{Sec.}{Secs.}
\Crefname{section}{Section}{Sections}
\Crefname{table}{Table}{Tables}
\crefname{table}{Tab.}{Tabs.}

\def\paperID{10524} 
\def\confName{CVPR}
\def\confYear{2024}

\begin{document}

\title{MRC-Net: 6-DoF Pose Estimation with MultiScale Residual Correlation}

\author{Yuelong Li \thanks{Equal contribution}\\
Amazon Inc.\\
{\tt\small yuell@amazon.com}
\and
Yafei Mao \footnotemark[1]\\
Amazon Inc.\\
{\tt\small yafeimao@amazon.com}
\and
Raja Bala\\
Amazon Inc.\\
{\tt\small rajabl@amazon.com}
\and
Sunil Hadap\\
Amazon Inc.\\
{\tt\small hadsunil@lab126.com}
}
\maketitle

\begin{abstract}
We propose a single-shot approach to determining 6-DoF pose of an object with available 3D \tcr{computer-aided design (CAD)} model from a single RGB image. Our method, dubbed MRC-Net, comprises two stages. The first performs pose classification and renders the 3D object in the classified pose. The second stage performs regression to predict fine-grained residual pose within class. Connecting the two stages is a novel multi-scale residual correlation (MRC) layer that captures high-and-low level correspondences between the input image and rendering from first stage. MRC-Net employs a Siamese network with shared weights between both stages to learn embeddings for input and rendered images. To mitigate ambiguity when predicting discrete pose class labels on symmetric objects, we use soft probabilistic labels to define pose class in the first stage. 
We demonstrate state-of-the-art accuracy, outperforming all competing RGB-based methods on four challenging BOP benchmark datasets: T-LESS, LM-O, YCB-V, and ITODD. Our method is non-iterative and requires no complex post-processing. Our code and pretrained models are available at \href{https://github.com/amzn/mrc-net-6d-pose}{https://github.com/amzn/mrc-net-6d-pose}.
\end{abstract}
\vspace{-0.2cm}

\section{Introduction}\label{sec:intro}

Estimating 3D object pose (rotation and translation) relative to the camera from a single image is a fundamental problem in many computer vision applications including robotics, autonomous navigation, and augmented reality. This task is challenging due to the complex shapes of real world objects, and diversity in object appearance due to lighting, surface color, background clutter, object symmetry, and occlusions.

A common solution is to directly regress object poses from images using deep neural networks~\cite{xiang2017posecnn, li2018deepim}. Alternatively the problem can be framed as one of classification, predicting a pose in terms of discrete buckets~\cite{kehl2017ssd, cai2022sc6d}. There have also been attempts to combine the two approaches, predicting a coarse pose class and then regressing residual pose within class~\cite{mousavian20173d}.  While residual regression helps reduce quantization errors from classification to some extent, performance generally falls short of state-of-the art, especially in challenging scenarios where there is lack of object texture or heavy occlusion. We believe a key reason is that in current approaches, the problem is formulated as multitask learning, where classification and regression tasks are trained \emph{in parallel} with shared top-level features. Such a design does not enable the regression task to receive direct guidance from the classification step. 

\begin{figure}[!t]
  \centering
  \includegraphics[width=\columnwidth]{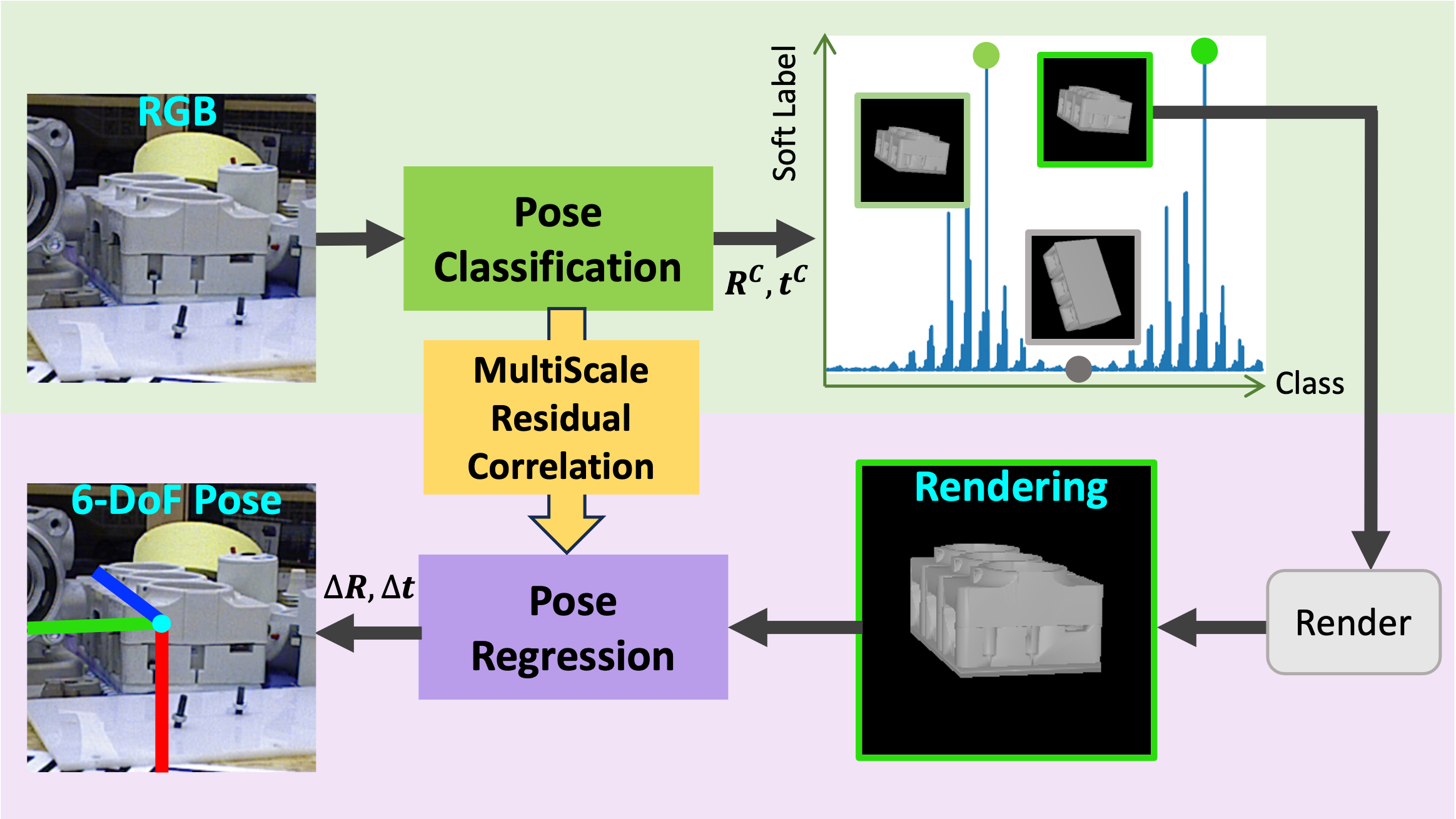}
  \caption{MRC-Net features a \emph{single-shot} sequential Siamese structure of two stages, where the second stage conditions on the first classification stage outcome through multi-scale residual correlation of poses between input and rendered images.}\label{fig:overview}
  \vspace{-0.3cm}
\end{figure}

We approach the problem differently. Since the tasks of classification and residual pose regression are inherently sequential, we hypothesize that it is more effective and natural to also learn them sequentially. We thus propose a two stage deep learning pipeline. In the first stage a classifier predicts pose in terms of a set of pose buckets. In the second stage, a deep regressor predicts residual pose within-bucket using features from the first stage. The classifier and regressor are implemented as Siamese networks with shared weights. Bridging the two stages is a novel multiscale residual correlation (MRC) layer  that draws on the well known render-and-compare paradigm to capture correspondences between the input pose and the pose rendered from the first classification stage. The multiscale architecture enables both local and global correspondences to inform pose estimation. To reduce ambiguity in discrete pose classification for objects with symmetry, we define class membership in terms of soft probabilistic labels. The overall network, dubbed MRC-Net is shown in Figure~\ref{fig:overview}. We validate our hypothesis in experiments, showing that the simple concept of sequential pose estimation with the MRC layer produces a major boost in performance and outperforms state-of-the-art techniques on the BOP Challenge~\cite{hodavn2020bop} datasets without the need for pre-initialization, iterative refinement, and post-processing. 

To summarize, the main contributions of this work are:
\begin{itemize}
    \vspace{0.2em}
    \item MRC-Net, a novel single-shot approach to directly estimate the 6-DoF pose of objects with known 3D models from monocular RGB images. Unlike prior methods, our approach performs classification and regression sequentially, guiding residual pose regression by conditioning on classification outputs. Moreover, we introduce a custom classification design based on soft labels, to mitigate symmetry-induced ambiguities.
    \vspace{0.2em}
   \item  A novel MRC layer that implicitly captures correspondences between input and rendered images at both global and local scales. Since MRC-Net is end-to-end trainable, this encourages the correlation features to be more discriminative, and avoids the need for complicated post-processing procedures.
       \vspace{0.2em}
   \item State-of-the-art accuracy on a variety of BOP benchmark datasets, advancing average recall by $2.4\%$ on average compared to results \tcr{reported by} competing methods.
       \vspace{0.2em}
\end{itemize}

\section{Related Work}\label{sec:related}

\tcr{Traditionally}, 6-DoF pose estimation has been solved via feature correspondences or template matching~\cite{collet2010efficient,hinterstoisser2011multimodal}. We focus our review on more recent learning-based methods relevant to our approach.

\textbf{Direct pose estimation} aims to determine 6-DoF pose in an end-to-end fashion without resorting to PnP solvers. Due to the continuous nature of object poses, the problem can be naturally formulated as one of regression ~\cite{xiang2017posecnn,li2019cdpn}. As object pose spans a large range, a popular approach is to iteratively refine an initial pose hypothesis through incremental updates~\cite{li2018deepim, labb2020cosypose}. Regression methods might get trapped in poor local minima unless properly initialized. To address this difficulty, some works opt for pose classification ~\cite{sundermeyer2018implicit,sundermeyer2020multi,cai2022sc6d, kehl2017ssd}. Although more robust and reliable at a coarse scale, classification suffers from loss of accuracy due to quantization of the pose continuum. To tackle this, some
methods add a parallel residual regression task ~\cite{mousavian20173d}. However, the regression branch is only indirectly connected to classification outcomes through a shared backbone. In contrast, our pipeline is sequential, where the residual regressor receives explicit guidance from the classifier via the MRC layer. To handle objects unseen during training, Megapose~\cite{labbe2022megapose} extends CosyPose~\cite{labb2020cosypose} by initializing pose via a coarse classifier, followed by a refiner. However, these two stages are independent networks that are separately trained. In contrast, our method strongly couples the two stages via a Siamese architecture and MRC layer, and is trained end-to-end. 

\textbf{Render-and-compare} is a strategy widely employed for
refining pose estimates by feeding additional rendered images alongside the input image into the deep network. Pioneering this research direction,
\cite{li2018deepim} and~\cite{manhardt2018deep} introduced
an iterative process that progressively renders an image based on the current pose estimate.
CosyPose~\cite{labb2020cosypose} generalizes this to a multi-view setting. CIR~\cite{lipson2022coupled},
built upon~\cite{teed2020raft}, estimates optical flow as a dense
2D-to-2D correspondence between real and rendered images and introduces a
differentiable PnP solver to perform coupled updates on both pose and
correspondence. In addition to 2D-to-2D correspondence,
RNNPose~\cite{xu2022rnnpose} optimize a 3D context encoder using Levernberg-Marquadt
optimization~\cite{nocedal1999numerical}. PFA~\cite{hu2022perspective} propose a non-iterative method by ensembling flow fields from the exemplars
towards the input image. SCFlow~\cite{hai2023shape} restricts indexing
correlation volume within the projected target's 3D shape to alleviate false
correspondences. All these methods estimate intermediate dense correspondences and require proper initialization using off-the-shelf networks. Our approach uses render-and-compare, but does not require pose initialization, and directly uses MRC features to predict pose in a single pass.

\textbf{Correspondence-based methods} use deep networks to predict an intermediate sparse or dense 3D-to-2D correspondence map, then predict pose through iterative PnP solvers. Earlier works focused on finding sparse corresponding geometric~\cite{rad2017bb8} or semantic ~\cite{peng2019pvnet,song2020hybridpose,lian2023checkerpose} keypoints with iterative refinement strategies~\cite{castro2023crt}. As sparse keypoints can be easily occluded, more recent techniques predict dense correspondences coupled with robust matching techniques ~\cite{li2019cdpn,hu2020single,shugurov2021dpodv2,SurfEmb}. Instead of per-pixel correspondences, some methods explore alternative surface representations for correspondence matching, such as recursive binary surface encoding~\cite{su2022zebrapose}, surface fragments~\cite{hodan2020epos}. Correspondence based methods involve computationally complex PnP solvers and are hard to train end-to-end, which can lead to suboptimal results.

\textbf{Other} works explore diverse avenues, such as knowledge distillation~\cite{guo2023knowledge}, new loss functions~\cite{liu2023linear}, integrating object detection~\cite{hai2023rigidity}, generalizing towards unseen objects~\cite{zhao2023learning}. Since depth data naturally aids 6-DoF pose estimation, several works~\cite{hodan2020epos, lipson2022coupled, su2022zebrapose, SurfEmb} explore this direction. While our method can be extended to incorporate depth, in this work we focus solely on single RGB image inputs to maintain broad scope and applicability. 



\section{Methodology}\label{sec:method}
\begin{figure*}[!t]
  \centering
  \includegraphics[width=0.85\textwidth]{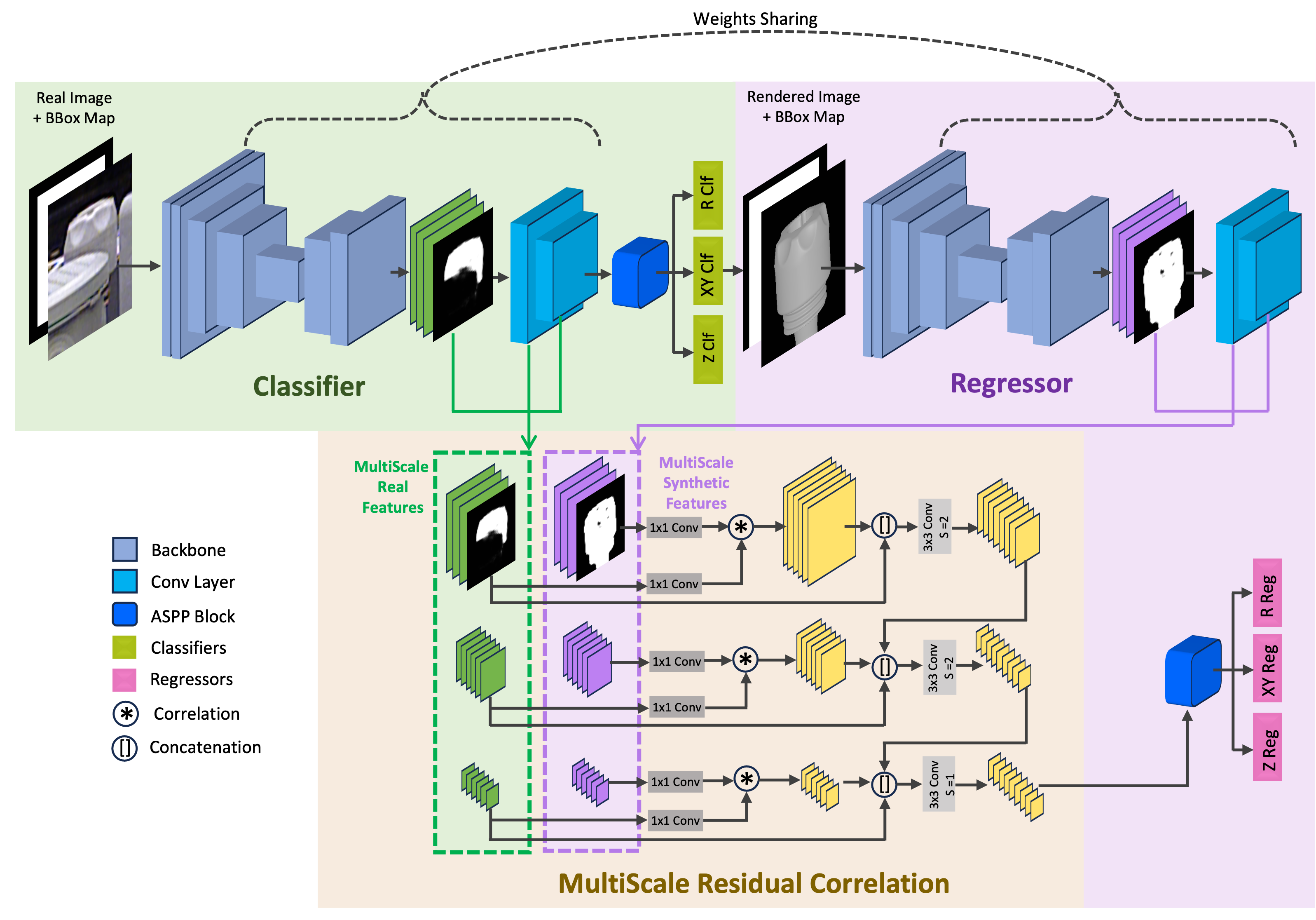}
  \caption{\textbf{MRC-Net Architecture.} The classifier and regressor stages employ a Siamese structure with shared weights. Both stages take the object crop and its bounding box map as input, and extract image features to detect the visible object mask, which are concatenated together to estimate object pose. The classifier first predicts pose labels. \tcr{These predictions, along with the 3D CAD model, are then used to render an image estimate, which serves as input for the second stage.} Features from the rendered image are correlated with those from real images in the MRC layer. These correlation features undergo ASPP processing within the rendered branch to regress the pose residuals.}\label{fig:network}\vspace{-1em}
\end{figure*}

\vspace{-0.1em}
Given a single RGB image crop (expanded and padded to square) around the object
of interest and the 3D model of the object, MRC-Net estimates 3D object rotation $\mathbf{R}$ and translation $\mathbf{t}$. In
practice, the image crop may be generated by an off-the-shelf object detector
such as Mask RCNN~\cite{he2017mask}. We concatenate the crop with the
binary encoding of the detection bounding box, dilated to model detection
inaccuracies, as input to the network. The architecture is shown in Figure~\ref{fig:network}.

The classification and residual regression stages solve three concurrent subtasks, namely, 3D rotation $\mathbf{R}\in SO(3)$,
2D in-plane translation $(t_x, t_y)\in\mathbb{R}^2$, and 1D depth $t_z\in\mathbb{R}$. We employ a Siamese architecture in both stages, incorporating the same asymmetrical ResNet34-UNet (with shared weights) to encode image features and discern the visible object mask. The mask is explicitly supervised during training. To capture local and global features, we use Atrous Spatial Pyramid Pooling (ASPP)~\cite{chen2017deeplab} modules before both classifier and regressor. 


In the first stage, we process the RGB input through a pose classifier that predicts pose labels $\mathbf{R}^c, (t_x^c,t_y^c,t_z^c)$, and renders an image with the predicted pose. Our aim is to achieve granular quantization to minimize pose residuals, thereby reducing the burden on the subsequent regression stage. To quantize rotation $SO(3)$, we uniformly generate $K$ rotation prototypes ${\{\mathbf{R}_k\}}_{k=1}^K$ using the method described in~\cite{yershova2010generating}. This allows us to uniformly partition $SO(3)$ into $K$ buckets through nearest neighbor assignment. We choose $K = 4608$ purely from practical consideration that it is the largest number we can fit into GPU memory. Similarly, for the spatial quantization of $t_x$, $t_y$, and $t_z$, we first use the Scale-invariant Translation Estimation (SITE) approach from~\cite{li2019cdpn} to transform translation into a well-defined estimation target denoted as $(\tau_x, \tau_y, \tau_z)$, and then generate spatial uniform grids in 2D and 1D, respectively. 

We feed the rendered image from the first stage and its bounding box into the second stage regressor which estimates pose residuals $\Delta \mathbf{R}, (\Delta \tau_x,\Delta \tau_y,\Delta \tau_z)$ between input and rendered object based on render-and-compare. Thus pose is guided by the classification stage, simplifying the regression task and eliminating the need for iterative refinement. Unlike traditional render-and-compare approaches that concatenate real and rendered images as network inputs~\cite{li2018deepim, lipson2022coupled}, we calculate correlations between these images in the MRC layer, detailed in the next section. Finally, the predictions from both stages are combined to determine 6-DoF pose: $\widehat{\mathbf{R}} = \Delta\mathbf{R}\mathbf{R}^c$, $\hat{\tau}_i=\tau_i^c+\Delta \tau_i, i\in\{x,y,z\}$.

\subsection{MultiScale Residual Correlation}\label{ssec:correlation}
MRC computes correlations between real and rendered image feature volumes at three scales. The correlated features are subsequently fed into the regression head to produce residual pose. As shown in Figure ~\ref{fig:network}, we progressively aggregate real features, correlation features, and downsampled features from the previous finer scale. This process is designed to incorporate multi-scale correlation and image context into the regressor. At the finest level, we also concatenate and input the object visibility mask, making subsequent layers occlusion-aware. We include a $1\times 1$ convolution before the correlation operation, utilizing shared weights for both real and synthetic branches, to obtain a linear projection of correlation features. When computing correlations, we adhere to the convention outlined in~\cite{sun2018pwc}, constraining the correlation within a local $P\times P$ image window. Mathematically, given a real feature volume $\mathbf{f}_r$ and a synthetic one $\mathbf{f}_s$, both with dimensions $d\times H\times W$, the resulting correlation volume $\mathbf{c}$ (of shape $P^2\times H\times W$) has elements defined by:
\[
  \mathbf{c}(\mathbf{v},\mathbf{x})=\frac{1}{\sqrt{d}}{\mathbf{f}_s(\mathbf{x})}^T\mathbf{f}_r(\mathbf{x}+\mathbf{v}),\qquad\forall\mathbf{v}:\|\mathbf{v}\|_\infty\leq P,
\]
where $\mathbf{x}$ indexes each pixel location and $\mathbf{v}$ indexes spatial
shift. Several techniques exist for formulating correlations that aim to capture
long range and higher order correspondences ~\cite{teed2020raft,dosovitskiy2015flownet}. Our approach is based on the observation that the correspondences are generally within short range thanks to our class conditional framework.

\begin{figure}[!t]
  \centering
  \begin{subfigure}[b]{0.8\linewidth}
    \includegraphics[width=\linewidth]{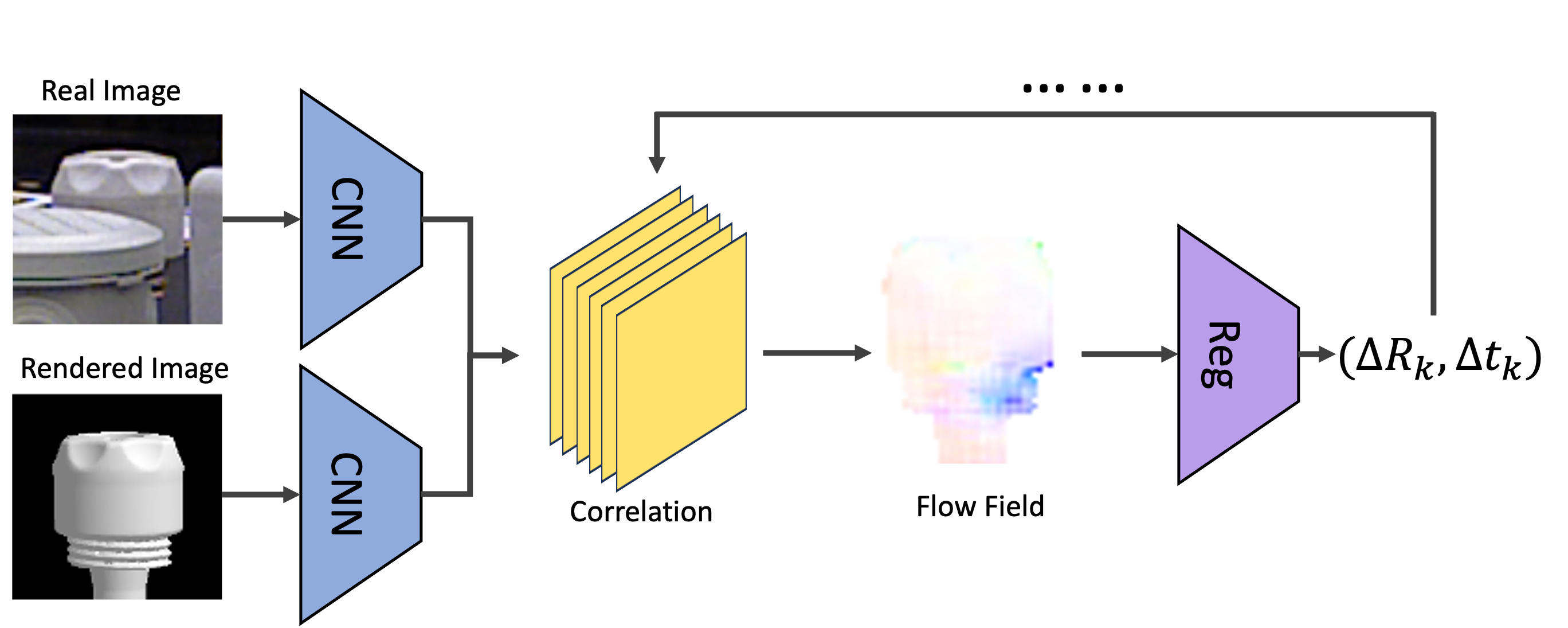}
    \caption{\label{subfig:corr_conventional}}
  \end{subfigure}
  \begin{subfigure}[b]{0.8\linewidth}
    \includegraphics[width=\linewidth]{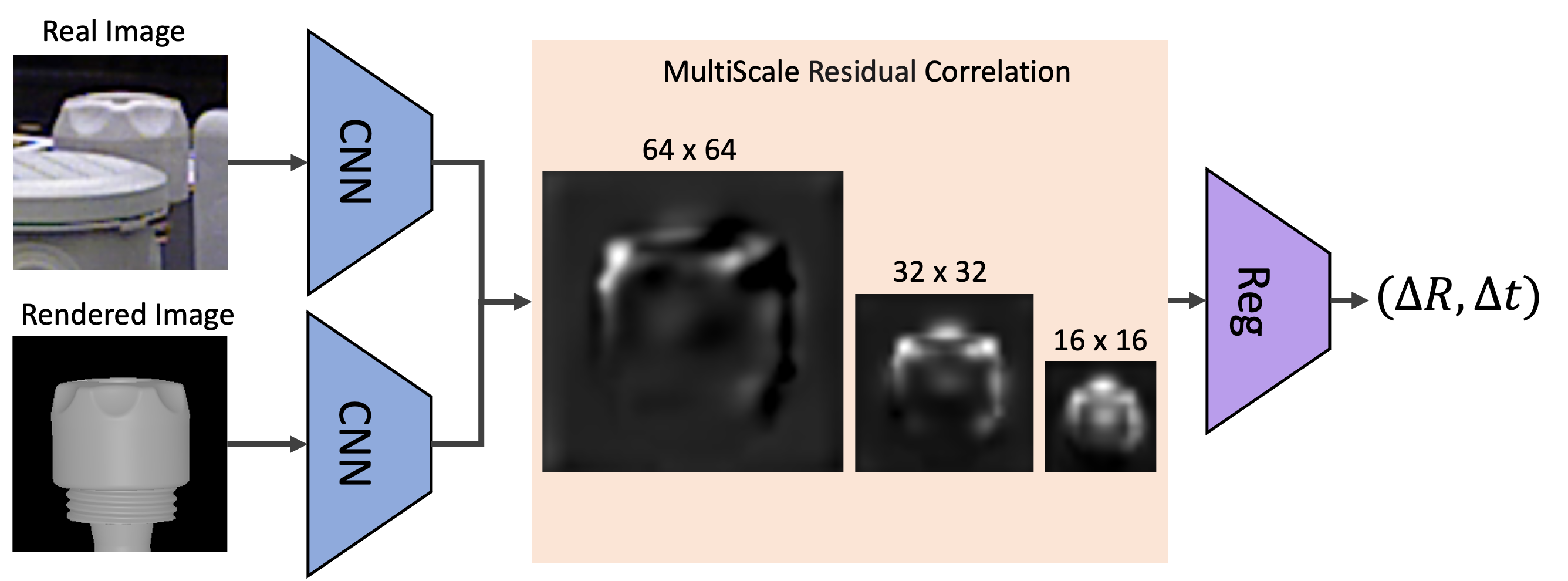}
    \caption{\label{subfig:corr_proposed}}
  \end{subfigure}
  \caption{Comparison of~\subref{subfig:corr_conventional} conventional approaches~\cite{lipson2022coupled} and~\subref{subfig:corr_proposed} our proposed approach leveraging feature correlations to estimate residual pose. Instead of predicting an intermediate flow field, we directly feed multi-scale feature correlations into the regression head in an end-to-end fashion. These features are discriminative and eliminate the need for post refinement, outlier removal or multiview renderings~\cite{lipson2022coupled, hai2023shape, hu2022perspective}.}\label{fig:correlations} 
\end{figure}

We contrast our render-and-compare approach with the common approach of computing optical flow to obtain correspondences between input and rendered images ~\cite{lipson2022coupled, hu2022perspective, hai2023shape} (Fig.~\ref{subfig:corr_conventional}). This flow field may capture inaccurate or redundant correspondences, requiring either an ensemble of flow fields from multiple exemplars \cite{hu2022perspective} or a recurrent coupled refinement of correspondence and pose \cite{lipson2022coupled, hai2023shape}. Moreover, most approaches involve non-differentiable operations like RANSAC and PnP~\cite{xu2022rnnpose,hu2022perspective}, making the system not end-to-end trainable. In contrast, our approach (Fig.~\ref{subfig:corr_proposed}) directly feeds multi-scale correspondence features to the regressor. This encourages the network to learn discriminative image features capturing correlations, and eliminates the need for an extra iterative/recurrent refinement of the correspondence field. Additionally, we explicitly feed the visibility mask to the correlation module to promote occlusion-robust learning. Our approach is end-to-end trainable. 

\subsection{Technical Details}\label{ssec:implementation}
\textbf{Soft labels for classification}: 
Assigning object rotations to discrete pose buckets presents a complex challenge. Specifically, object rotations may reside on decision boundaries of $SO(3)$ buckets, having equal geodesic distances towards multiple prototypes, introducing inherent ambiguity. Additionally, object symmetry can introduce invariance under certain rigid transformations, resulting in multiple valid poses. To address this uncertainty, we employ \emph{soft labels} for the classification task, modeling the \emph{binary} probabilities of whether or not the objects belong to each view bucket based on pose error metrics. A similar concept is explored in~\cite{chen2022epro}, where a continuous distribution of object poses is defined based on reprojection error. In contrast, we formulate soft assignments for rotation classes, offering a novel perspective to the problem.

In detail, given the annotated object pose $(\mathbf{R}^\ast,\mathbf{t}^\ast)$, we define the rotation labels as:
\begin{equation}
  \begin{split}
    l^\mathbf{R}_k&=\exp\left\{-\frac{\rho_{\mathrm{pose\_symm}}(\mathbf{R}^\ast,\mathbf{t}^\ast;\mathbf{R}_k,\mathbf{t}^\ast)}{\sigma}\right\},
  \end{split}\label{eq:rot_label}
\end{equation}
where $k=1,\dots,K$, $\rho_{\mathrm{pose\_symm}}(\mathbf{R}_1,\mathbf{t}_1;\mathbf{R}_2,\mathbf{t}_2)$
measures the symmetry-aware distance between object poses
$(\mathbf{R}_1,\mathbf{t}_1)$ and $(\mathbf{R}_2,\mathbf{t}_2)$~\cite{labb2020cosypose}, and $\sigma>0$ is a hyperparameter regulating the concentration of classification labels. Throughout our experiments, we fix $\sigma=0.03\,d_{\mathrm{diam}}$ where $d_{\mathrm{diam}}$ is the object diameter \tcr{defined as the farthest pairwise vertex distances}. Essentially, soft labels depict a exponentially-weighted combination of prototype rotations $\mathbf{R}_k$ centered around the annotated rotation $\mathbf{R}^\ast$. The weighting is determined by their symmetry-aware distance.

Similarly, to generate soft labels for translations, we uniformly quantize $(\tau_x,\tau_y)$ into a $64\times64$ grid, and $\tau_z$ into $1000$ bins within its specified range. The translation labels are finally computed using Gaussian functions centered around the ground truth location.

\textbf{Loss functions}: The rotation classification task is
trained by minimizing the focal loss $\mathcal{L}^\mathbf{R}_{\mathrm{cls}}$~\cite{lin2017focal}:
\vspace{-0.2em}
\begin{equation*}
  \mathcal{L}^\mathbf{R}_{\mathrm{cls}}=\sum_{k=1}^K -w^+l^\mathbf{R}_k(1-\hat{p}_k)^2\log\hat{p}_k - (1 - l^\mathbf{R}_k)\hat{p}_k^2\log(1 - \hat{p}_k),
\end{equation*}
\vspace{-0.2em}
where ${\{\hat{p}_k\}}_{k=1}^K$ are probabilities predicted by the classifier and $l_k^\mathbf{R}$'s are defined in~\eqref{eq:rot_label}. Note that we apply pairwise binary encoding to allow for multiple class affiliations simultaneously and introduce a weighting parameter $w^+ > 0$, which we fix to $100$, to address class imbalances.

To solve for translation, both $xy$ and $z$ classification tasks are optimized on multi-class focal loss:
\[\mathcal{L}^i_{\mathrm{cls}}=-{\left(1-\sum_{j=1}^{n_i}{l_j^i}\hat{p}_j^i\right)}^2\log\left(\sum_{j=1}^{n_i}l_j^i \hat{p}_j^i\right),\quad i\in\{xy,z\},\]
where $l_j^{xy}$ and $l_j^z$ are the translation soft labels for $xy$ and $z$, $\hat{p}_j^{xy}$ and $\hat{p}_j^z$ are predicted probabilities, $n_{xy}=64^2$ and $n_z=1000$. At inference time, we pick the class with highest confidence for all three tasks.

The rotation regression task predicts residual rotation $\Delta\mathbf{R}$ between rendered and real object. This
quantity is typically localized with smaller pose angles, and easier to learn via regression. Combining the coarse rotation $\mathbf{R}^c$ from the classifier with $\Delta \mathbf{R}$ yields the fine-grained rotation prediction $\hat{\mathbf{R}}=\Delta\mathbf{R} \mathbf{R}^c$. We adopt the 6D rotation representation suggested by~\cite{zhou2019continuity} as it generally yields better accuracy and reduces the occurrence of large errors. Similarly, we combine  translation estimates from the two stages to arrive at the final object translation: $\hat{\tau}_i=\tau_i^c+\Delta\tau_i,\quad i\in\{x,y,z\}$


To train the rotation regression task, disentangled loss~\cite{simonelli2019disentangling} is used
\[
\mathcal{L}_{\mathrm{reg}}^\mathbf{R}=\rho_{\mathrm{pose\_sym}}(\Delta\mathbf{R}\mathbf{R}^c,\mathbf{t}^\ast;\mathbf{R}^\ast,\mathbf{t}^\ast),
\]
where $(\mathbf{R}^\ast,\mathbf{t}^\ast)$ is the annotated object pose. The other two terms $\mathcal{L}_{\mathrm{reg}}^{xy}$ and $\mathcal{L}_{\mathrm{reg}}^z$ are defined similarly. The final loss function is formed via a weighted combination of the individual terms:
\begin{equation}
  \begin{split}
    &\mathcal{L}=w_{\mathrm{cls}}^\mathbf{R}\mathcal{L}_{\mathrm{cls}}^\mathbf{R}+w_{\mathrm{cls}}^{xy}\mathcal{L}_{\mathrm{cls}}^{xy}+w_{\mathrm{cls}}^z\mathcal{L}_{\mathrm{cls}}^z\\
    &+w_{\mathrm{reg}}^\mathbf{R}\mathcal{L}_{\mathrm{reg}}^\mathbf{R}+w_{\mathrm{reg}}^{xy}\mathcal{L}_{\mathrm{reg}}^{xy}+w_{\mathrm{reg}}^z\mathcal{L}_{\mathrm{reg}}^z+w^M\mathcal{L}^M,
  \end{split}\label{eq:loss}
\end{equation}
where \tcr{$\mathcal{L}^M$ is the visible mask term defined as the binary cross entropy loss and $w^M>0$ is its weight.}

\textbf{Perspective correction}: As our method operates on a cropped view centered around the object of interest, it may lack global context. Specifically, as highlighted in~\cite{li2022cliff}, situations may arise where the network is compelled to predict different poses for image crops with identical appearances, causing confusion in supervision. A common strategy to mitigate this challenge involves converting egocentric rotations into allocentric ones~\cite{wang2021gdr,cai2022sc6d}. However, this solution demands accurate prediction of the object center and does not address translation considerations. Instead, we additionally incorporate the global information of the bounding box features into each classifier, similar to~\cite{li2022cliff}. We feed $\left[\frac{b_x - c_x}{f}, \frac{b_y - c_y}{f}, \frac{s_{bbox}}{f}\right]$, where $b_x$, $b_y$, $c_x$, $c_y$, and $f$ represent the bounding box center, camera principal point, and focal length, respectively. In our experiments, as a common practice we directly retrieve the camera focal length and principal point from dataset annotations. Different from ~\cite{li2022cliff}, we obtain the coarse translation estimate $({t}^c_x, {t}^c_y, {t}^c_z)$ after the classification stage. Subsequently, in the regression render-and-compare stage, we directly feed $\left[\frac{{t}^c_x}{{t}^c_z}, \frac{{t}^c_y}{{t}^c_z}, \frac{s_{bbox}}{f}\right]$ to the regression head instead of the bounding box center. This modification ensures that the regressor is informed about the true projected object center, thereby promoting accurate prediction of the residuals.

\section{Experiments}\label{sec:experiments}
We systematically analyze the novel components of our method and perform comparisons with state of the art. 

\subsection{Setup}\label{ssec:settings}
We use AdamW optimizer~\cite{loshchilov2018decoupled} with cosine annealing
learning rate scheduler to train our network. The learning rate is initially set to $2e^{-4}$ and
gradually reduced to $5e^{-6}$ \tcr{with a batch size of 128}. We render images using PyTorch3D~\cite{ravi2020pytorch3d} \tcr{with constant directional lights along the optical axis, without texturing or antialising}. We warm-up the learning rate linearly in the first 1K
iterations. We choose $w_{\mathrm{cls}}^\mathbf{R}=0.05,w_{\mathrm{cls}}^{xy}=2.0,w_{\mathrm{cls}}^z=2.0,w_{\mathrm{reg}}^\mathbf{R}=1.0,w_{\mathrm{reg}}^{xy}=1.0,w_{\mathrm{reg}}^z=0.2$ and $w^M=10.0$ to balance individual loss terms in~\eqref{eq:loss}. These parameters are fixed for all experiments.

\tcr{Following the BOP Challenge 2022 standard, we employ Mask RCNN detections from~\cite{labb2020cosypose} to extract image crops.} We evaluate our work on four challenging and widely cited BOP benchmark datasets: T-LESS~\cite{hodan2017tless}, ITODD~\cite{drost2017introducing}, YCB-V~\cite{xiang2017posecnn}, and LM-O~\cite{brachmann2014learning}. These cover a large variety of difficult cases such as object symmetries, occlusions, absence of textures, etc. A detailed description of these datasets is available on the BOP Challenge website~\footnote{\label{bopnote}https://bop.felk.cvut.cz/challenges/}. For each dataset, our model is pre-trained on synthetic images
generated with physically-based rendering (PBR), with a batch size of 128 for a
total of 75 epochs on 8 NVIDIA V100 GPUs. We then fine-tune our model using a
mixture of real and synthetic images for T-LESS and YCB-V datasets. We train one single model per dataset. 
The official BOP Challenge website\textsuperscript{\ref{bopnote}} does not provide real images in the training set for ITODD and LM-O; hence, we include results trained exclusively on the synthetic images for these two datasets.

We use the same basic backbone (ResNet34) as many existing works~\cite{wang2021gdr,cai2022sc6d,su2022zebrapose} rather than opting for heavier versions for performance.
We apply domain randomization techniques similar to~\cite{wang2021gdr} to
reduce overfitting caused by the domain gap between real and synthetic data. At
inference time, we employ the same test-time augmentation (TTA) technique as~\cite{SurfEmb,cai2022sc6d}: we
rotate the input image crop by $90^\circ$, $180^\circ$, $270^\circ$ and
$360^\circ$, run inference on these four copies independently, and choose the
one with the highest classification score as the final pose prediction.

We adhere to the latest BOP evaluation protocol~\cite{hodavn2020bop}. Three error metrics are calculated on each object per image: Visible Surface Discrepancy (VSD), Maximum Symmetry-Aware Surface Distance (MSSD), and Maximum Symmetry-Aware Projection Distance (MSPD). An average recall is then computed for each metric by aggregating recall rates at different pre-determined thresholds, to obtain per-metric recalls ($\text{AR}_\text{VSD}, \text{AR}_\text{MSSD}$, and $\text{AR}_\text{MSPD}$). Finally, these are averaged to obtain an overall Average Recall (AR). Detailed definitions of these metrics can be found in~\cite{hodavn2020bop}. Some approaches only report results in ADD(-S), AUC of ADD-S, and AUC of ADD(-S). We show comparisons of these methods on the YCB-V dataset in these metrics, with detailed definitions available in~\cite{xiang2017posecnn}.

\subsection{Ablation Studies}\label{ssec:ablation}
We explore the significance of each novel component of our design on the challenging T-LESS dataset. The investigation is based solely on using synthetic PBR images for training. Additional ablation studies can be found in the supplementary material.

\textbf{Hard label vs soft labels:} 
We hypothesize that soft labels enable a more nuanced representation, especially in scenarios where a unique class label is hard to identify due to object symmetries or equiprobable pose hypotheses. To verify its efficacy, 
we compare training with soft and hard labels, where for the latter, we pick a unique class index with the highest soft label value. Results are summarized in Table~\ref{tab:classify}. While hard labels are able to maintain reasonable classification performance, soft labels achieve an increase of \tcr{2.3\% in average AR}, clearly validating its effectiveness.

\textbf{Classification and regression in parallel vs sequential:} \tcr{We start by designing a classification-only baseline model that comprises only the first stage of MRC-Net. Then, we extend it into a parallel (multitask) architecture by adding an additional regression head atop the pooled feature layer.}
Next, we implement a simple sequential pipeline that concatenates real and rendered features in $64\times 64$ resolution from stage 1 and passes these to the stage 2 regressor. No MRC layer is included. AR metrics of these models are listed in the first \tcr{three} rows of Table~\ref{tab:abl_correlation}. Clearly, the parallel method leads to significant performance drop (11.3\% in average AR), proving the benefit of sequential class-conditioned regression.

An interesting observation is that simple addition of a parallel regression head (second row in Table~\ref{tab:abl_correlation}) offers little or no improvement over the classification-only baseline (first row). 
As the regression head is unaware of the classification outcome, there is little opportunity for it to correct for classification errors.

\textbf{MRC:} To quantify the benefits of the feature correlation block, we conduct experiments (Table~\ref{tab:abl_correlation}) on single-scale feature correlation, i.e., we only correlate the top-level features of $16\times 16$ resolution. This yields 0.3\% additional improvement of AR compared to simple concatenation, implying that correlation features are more discriminative to learn the residual pose. Extending to multiscale correlation further increases AR by 0.6\%, confirming that correlation at multiple scales provides complementary information.

\textbf{Perspective correction:} In Table~\ref{tab:abl_design}, we examine the impact of perspective correction (PerspCrrct) and TTA compared to the complete model. While not being a predominant factor, employing TTA still yields a plausible performance boost of $0.4\%$ AR which demonstrates its effectiveness. Similarly, the inclusion of perspective correction noticeably enhances performance by 0.5\%.

\begin{table}[h!]
  \begin{subtable}{\linewidth}
    \centering
    \setlength{\tabcolsep}{4pt}
    \begin{tabular}{c|ccc|c}
      \toprule
  Method & $\mathrm{AR_{VSD}}\!\!\uparrow$ & $\mathrm{AR_{MSSD}}\!\!\uparrow$ & $\mathrm{AR_{MSPD}}\!\!\uparrow$ & $\mathrm{AR}\!\!\uparrow$ \\
  \midrule
      Hard Label& $67.8$ & $72.2$ & $84.6$ & $74.8$ \\
      Soft Label & $70.6$ & $74.7$ & $86.0$ & $77.1$ \\
      \bottomrule
    \end{tabular}
    \vspace{0.4em}
    \caption{\tcr{\textbf{Comparison of hard and soft labels}. The table illustrates performance when trained with and without soft labels.}}\label{tab:classify}
    \vspace{0.8em}

    \setlength{\tabcolsep}{2.5pt}
    \begin{tabular}{c|c|ccc|c}
      \toprule
  Type & Method & $\mathrm{AR_{VSD}}\!\!\uparrow$ & $\mathrm{AR_{MSSD}}\!\!\uparrow$ & $\mathrm{AR_{MSPD}}\!\!\uparrow$ & $\mathrm{AR}\!\!\uparrow$ \\
  \midrule
      - & ClfOnly & $55.6$ & $62.8$ & $77.4$ & $65.3$ \\
      P & MultiTask & $55.3$ & $62.7$ & $76.6$ & $64.9$ \\
      S & FeatConcat & $69.4$ & $73.5$ & $85.7$ & $76.2$ \\
      S & SSCorr & $69.8$ & $73.9$ & $85.8$ & $76.5$ \\
      S & MRC-Net & $70.6$ & $74.7$ & $86.0$ & $77.1$\\
      \bottomrule
    \end{tabular}
    \vspace{0.4em}
    \caption{\textbf{Comparison of parallel and sequential designs.} \tcr{\textbf{ClfOnly} denotes a classification-only model.} \textbf{MultiTask} is a parallel baseline architecture. \textbf{FeatConcat} sends concatenated real and rendered image features to the regressor. \textbf{SSCorr} use single-scale feature correlation, while \textbf{MRC-Net} uses multi-scale feature correlation. P denotes parallel classification and regression, while S denotes sequential.}\label{tab:abl_correlation}
    \vspace{0.8em}

    \setlength{\tabcolsep}{3pt}
    \begin{tabular}{c|ccc|c}
      \toprule
  Method & $\mathrm{AR_{VSD}}\!\!\uparrow$ & $\mathrm{AR_{MSSD}}\!\!\uparrow$ & $\mathrm{AR_{MSPD}}\!\!\uparrow$ & $\mathrm{AR}\!\!\uparrow$ \\
  \midrule
      \makecell{w/o PerspCrrct} & $69.9$ & $74.1$ & $85.9$ & $76.6$\\
      \makecell{w/o TTA} & $70.4$ & $74.1$ & $85.7$ & $76.7$ \\
      Full model & $70.6$ & $74.7$ & $86.0$ & $77.1$\\
      \bottomrule
    \end{tabular}
    \vspace{0.4em}
    \caption{\textbf{Impact of perspective correction (PerspCrrct) and test-time augmentation (TTA).} Dropping perspective correction results in noticeable degradation in performance, while omitting TTA has a more modest impact.}\label{tab:abl_design}
  \end{subtable}
  \caption{Ablation studies of our method on T-LESS dataset~\cite{hodan2017tless}.}\label{tab:ablations}
\end{table}

\subsection{Comparison with State-of-the Art}\label{ssec:comparison}

\begin{table}[h!]
  \begin{subtable}{\columnwidth}
    \centering
    \setlength{\tabcolsep}{4pt}
    \begin{tabularx}{\linewidth}{c|ccccc}
      \toprule
 Method & {\small T-LESS} & {\small ITODD} & {\small YCB-V} & {\small LM-O} & {\small Avg.}\\
\midrule
EPOS~\cite{hodan2020epos} & 46.7 & 18.6 & 49.9 & 54.7 & 42.5\\
CDPNv2~\cite{li2019cdpn} & 40.7 & 10.2 & 39.0 & 62.4 & 38.1\\
DPODv2~\cite{shugurov2021dpodv2}  & 63.6 & - & - & 58.4 & -\\
PVNet~\cite{peng2019pvnet} & - & - & - & 57.5 & -\\
CosyPose~\cite{labb2020cosypose}  & 64.0 & 21.6 & 57.4 & 63.3 & 51.6\\
SurfEmb~\cite{SurfEmb}  & \underline{74.1} & \underline{38.7} & 65.3 & 65.6 & \underline{60.9}\\
SC6D~\cite{cai2022sc6d} & 73.9 & 30.3 & 61.0 & - & -\\
SCFlow~\cite{hai2023shape} & - & - & 65.1 & \underline{68.2} & - \\
PFA~\cite{hu2022perspective}   & - & - & 61.5 & 67.4 & -\\
CIR~\cite{lipson2022coupled}   & - & - & - & 65.5 & -\\
SO-Pose~\cite{di2021so}   & - & - & - & 61.3 & -\\
NCF~\cite{huang2022neural}   & - & - & \underline{67.3} & 63.2 & -\\
CRT-6D~\cite{castro2023crt}   & - & - & - & 66.0 & -\\
MRC-Net & \textbf{77.1} & \textbf{39.3} & \textbf{68.1} & \textbf{68.5}  & \textbf{63.3}\\
      \bottomrule
    \end{tabularx}
    \caption{}\label{subtab:comp_pbr}
  \end{subtable}
  \begin{subtable}{\columnwidth}
    \centering
    \begin{tabular}{c|ccc}
\toprule
     Method & T-LESS & YCB-V & Avg. \\
\midrule
CDPNv2~\cite{li2019cdpn}  & 47.8 & 53.2 & 50.5\\
CosyPose~\cite{labb2020cosypose} & 72.8 & 82.1 & 77.4\\
SurfEmb~\cite{SurfEmb} & 77.0 & 71.8 & 74.4\\
SC6D~\cite{cai2022sc6d} & \underline{78.0} & 78.8 & \underline{78.4}\\
SCFlow~\cite{hai2023shape} & - & \textbf{82.6} & - \\
CIR~\cite{lipson2022coupled} & 71.5 & \underline{82.4} & 77.0 \\
SO-Pose~\cite{di2021so}& - & 71.5 & -\\
NCF~\cite{huang2022neural}& - & 77.5 & -\\
CRT-6D~\cite{castro2023crt}   & - & 75.2 & -\\
MRC-Net & \textbf{79.8} & 81.7 & \textbf{80.8}\\
 \bottomrule
\end{tabular}
\caption{}\label{subtab:comp_real}
  \end{subtable}
  \caption{\textbf{Comparison with state-of-the-srt RGB methods on the BOP benchmarks.} We report Average Recall in \% on (a) \textbf{T-LESS, ITODD, YCB-V, and LM-O} datasets trained with purely \textbf{synthetic} images, and (b) \textbf{T-LESS and YCB-V} datasets also trained with \textbf{real} images. We highlight the best results in \textbf{bold} and \underline{underline} the second best results. \textquote{-} denotes results missing from the original paper.}\vspace{-1em}
  \label{tab:ar-comp}
\end{table}

\begin{table}[h!]
    \centering
    \setlength{\tabcolsep}{3.5pt}
    \begin{tabularx}{\linewidth}{c|ccc}
      \toprule
     Method & \makecell{ADD(-S)\\$\uparrow$} & \makecell{ AUC\\ADD-S $\uparrow$} & \makecell{AUC\\ADD(-S) $\uparrow$}  \\
\midrule
SegDriven~\cite{hu2019segmentation} & 39.0 & - & - \\
SingleStage~\cite{hu2020single} & 53.9 & - & - \\
CosyPose~\cite{labb2020cosypose} & - & 89.8 & 84.5 \\
RePose~\cite{iwase2021repose} & 62.1 & 88.5 & 82.0 \\
GDR-Net~\cite{wang2021gdr}  & 60.1 & 91.6 & 84.4 \\
SO-Pose~\cite{di2021so} & 56.8 & 90.9 & 83.9 \\
ZebraPose~\cite{su2022zebrapose}$^\ast$ & 80.5 & 90.1 & 85.3 \\
SCFlow~\cite{hai2023shape} & 70.5 & - & -\\
DProST~\cite{park2022dprost} & 65.1 & - & 77.4 \\
CheckerPose~\cite{lian2023checkerpose}$^\ast$ & \underline{81.4} & 91.3 & 86.4 \\
MRC-Net & 81.2 & \underline{95.0} & \underline{92.3} \\
\tcr{MRC-Net$^\ast$} & \tcr{\textbf{83.6}} & \tcr{\textbf{97.0}} & \tcr{\textbf{94.3}} \\
      \bottomrule
    \end{tabularx}

  \caption{\textbf{Comparison on the YCB-V Dataset.} We report the ADD(-S), AUC of ADD-S, and AUC of ADD(-S) metrics in \%. We highlight the best results in \textbf{bold} and \underline{underline} the second best results. \textquote{-} denotes results missing from the original paper \tcr{and $^\ast$ denotes results obtained using the FCOS detector~\cite{li2019cdpn}.}}\vspace{-0.7em}
  \label{tab:adds-comp}
\end{table}

\textbf{Quantitative comparisons}. We benchmark our method against state-of-the-art techniques spanning a variety of recent approaches:  EPOS~\cite{hodan2020epos}, CDPNv2~\cite{li2019cdpn}, DPODv2~\cite{shugurov2021dpodv2}, PVNet~\cite{peng2019pvnet}, CosyPose~\cite{labb2020cosypose}, SurfEmb~\cite{SurfEmb}, SC6D~\cite{cai2022sc6d}, SCFlow~\cite{hai2023shape}, CIR~\cite{lipson2022coupled}, PFA~\cite{hu2022perspective}, SO-Pose~\cite{di2021so}, NCF~\cite{huang2022neural}, CRT-6D~\cite{castro2023crt}, GDR-Net~\cite{wang2021gdr}, ZebraPose~\cite{su2022zebrapose}, DProST~\cite{park2022dprost}, RePose~\cite{iwase2021repose}, SegDriven~\cite{hu2019segmentation}, SingleStage~\cite{hu2020single}, and CheckerPose~\cite{lian2023checkerpose}. We report metrics from the original references.

For all four datasets, we present the $\text{AR}$ results using PBR training images. In addition, since real training images are available for T-LESS and YCB-V, we include $\text{AR}$ results after fine-tuning on these real images, adhering to the standard BOP protocol. Results are summarized in Table~\ref{tab:ar-comp}. From Table~\ref{subtab:comp_pbr}, it can be seen that our method achieves state-of-the-art performance across all datasets when trained on pure synthetic data. We outperform other models on ITODD and YCB-V datasets by a margin of 0.6\%, and 0.8\% respectively. Particularly on T-LESS, we significantly outperform others by 3\%. Surprisingly, our performance on pure synthetic training is even comparable to top-performing methods trained with real data in Table~\ref{subtab:comp_real}. This suggests that our method has strong potential in applications with limited real data, for example underwater~\cite{risholm2021underwater} and aerospace~\cite{hu2021wide}. On the other hand, when comparing fine-tuned results over YCB-V dataset, our performance lags behind the top performers by up to 0.9\%. Indeed, this dataset carries rich textures on the objects' surface, facilitating dense correspondence finding and iterative refinement among top-performing methods~\cite{labb2020cosypose,lipson2022coupled,hai2023shape}. Moreover, as discussed in~\cite{SurfEmb}, there are inaccurate CAD models and noisy ground truth annotations within this dataset.

Table~\ref{tab:adds-comp} shows non-BOP metrics (ADD(-S), AUC of ADD-S, and AUC of ADD(-S)) on the YCB-V dataset for methods~\cite{hu2019segmentation,labb2020cosypose, hu2020single, iwase2021repose, wang2021gdr, di2021so,su2022zebrapose, hai2023shape, park2022dprost, lian2023checkerpose} reporting these metrics. \tcr{These methods commonly rely on detections from~\cite{labb2020cosypose} or~\cite{li2019cdpn} for their evaluations. Therefore, we present results for both cases. More details are available in the supp. materials. Notably, MRC-Net offers a clear improvement of 2.2\%, 5.4\%, and 7.9\% in the three metrics.}

\begin{figure*}[h!]
\centering
\begin{subfigure}[b]{0.16\textwidth}
  \includegraphics[width=\columnwidth]{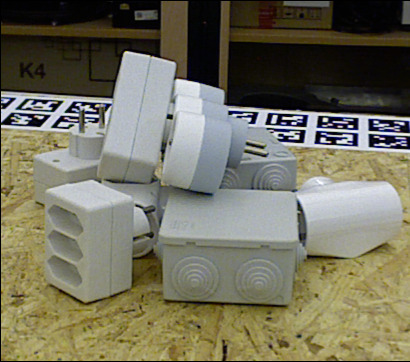}
\end{subfigure}
\begin{subfigure}[b]{0.16\textwidth}
  \includegraphics[width=\columnwidth]{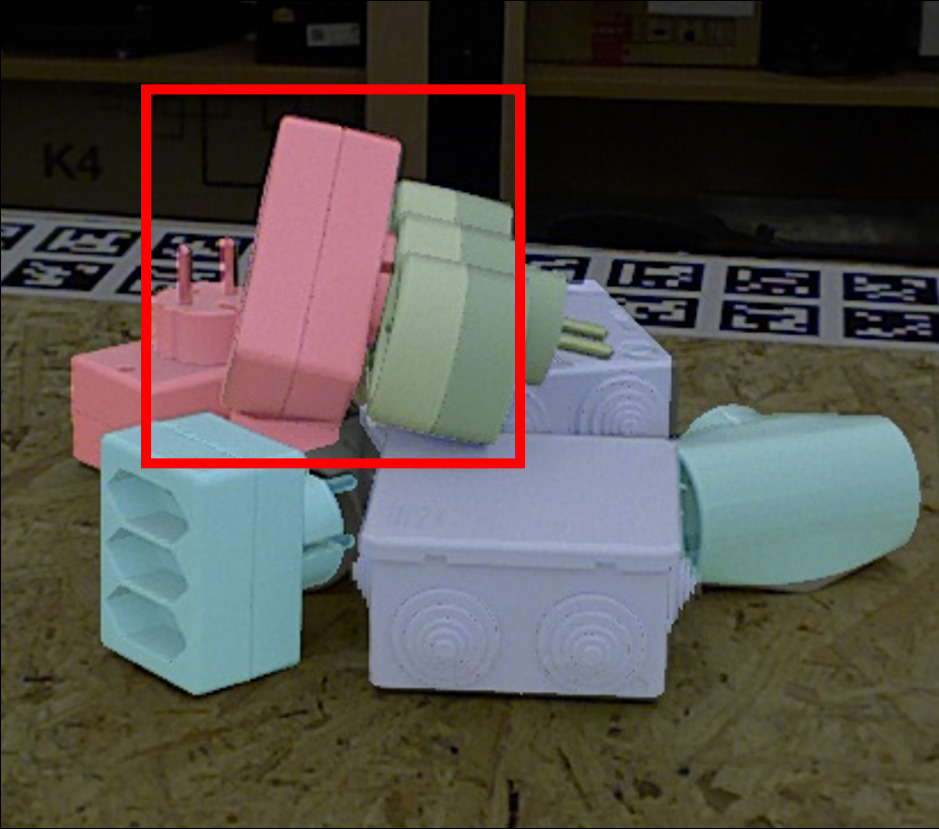}  
\end{subfigure}
\begin{subfigure}[b]{0.16\textwidth}
  \includegraphics[width=\columnwidth]{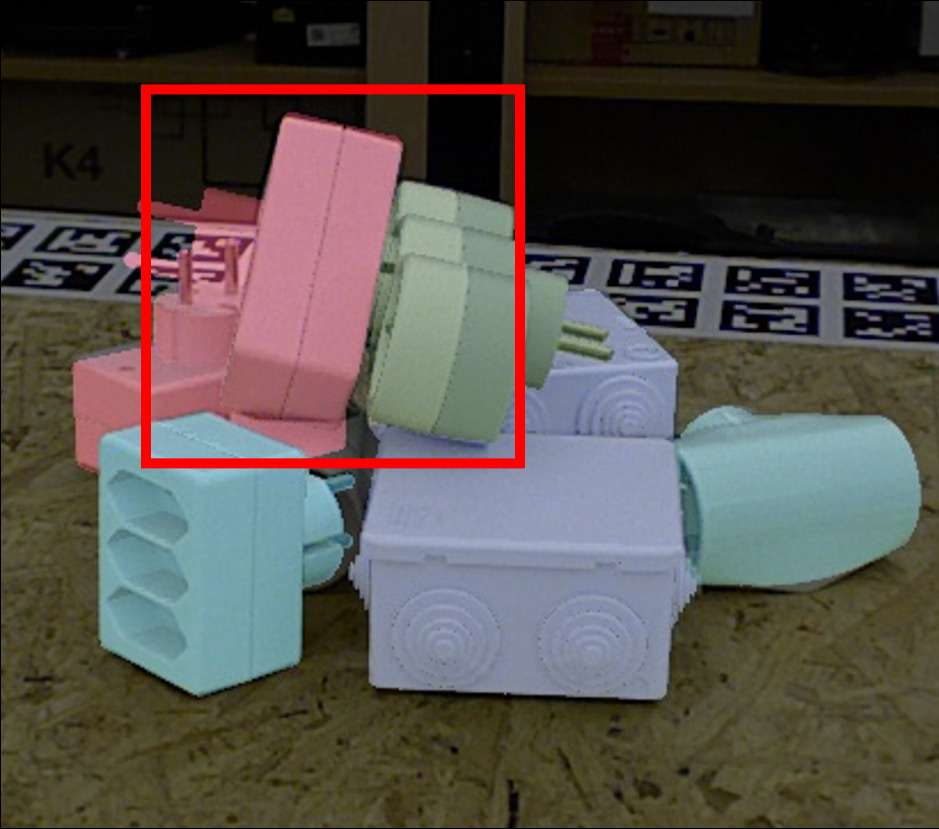}
\end{subfigure}
\begin{subfigure}[b]{0.16\textwidth}
  \includegraphics[width=\columnwidth]{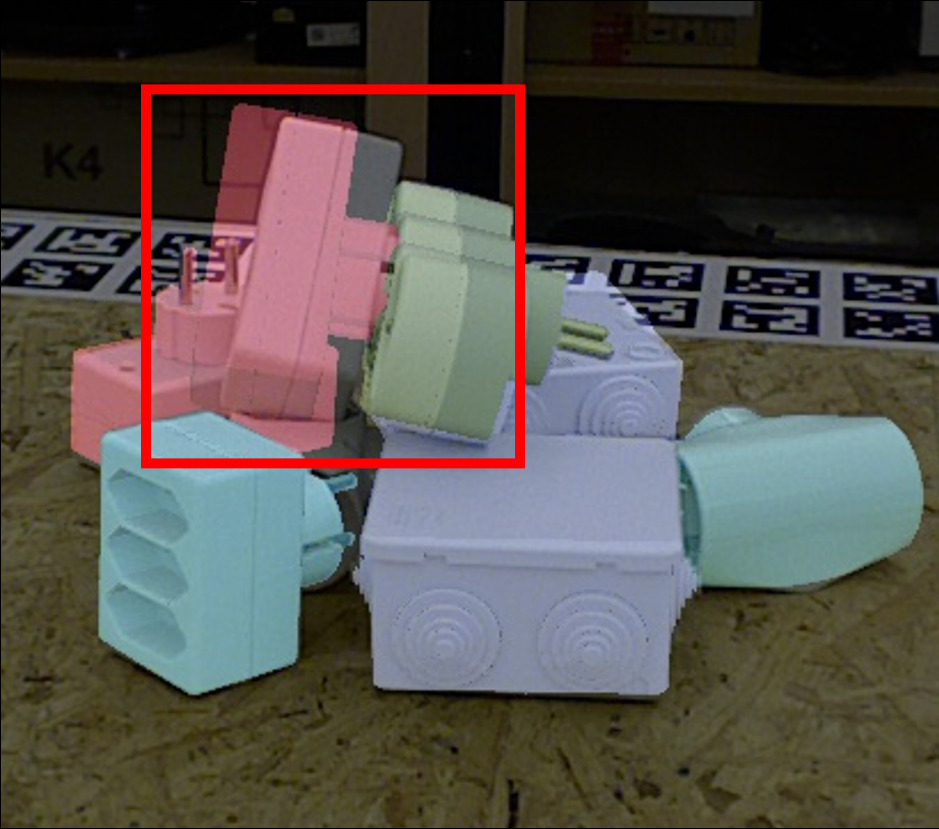}
\end{subfigure}
\begin{subfigure}[b]{0.16\textwidth}
  \includegraphics[width=\columnwidth]{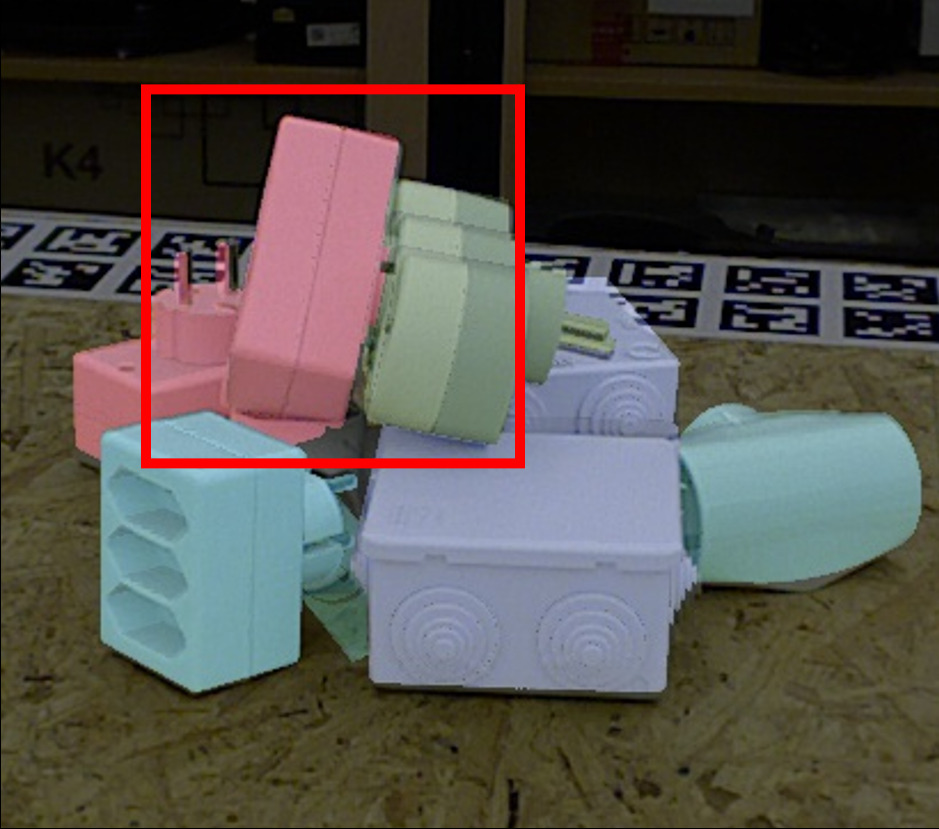}
\end{subfigure}

\begin{subfigure}[b]{0.16\textwidth}
  \includegraphics[width=\columnwidth]{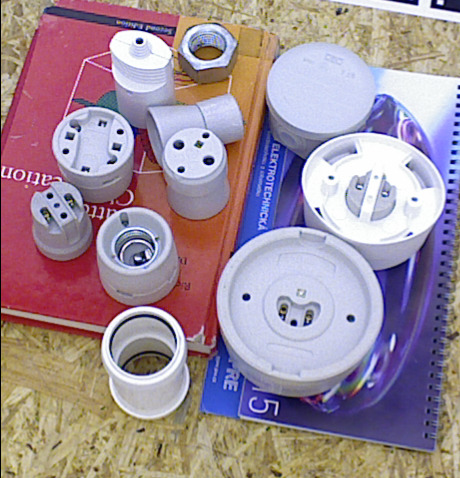}
  \caption{\label{subfig: overlaya}}
\end{subfigure}
\begin{subfigure}[b]{0.16\textwidth}
  \includegraphics[width=\columnwidth]{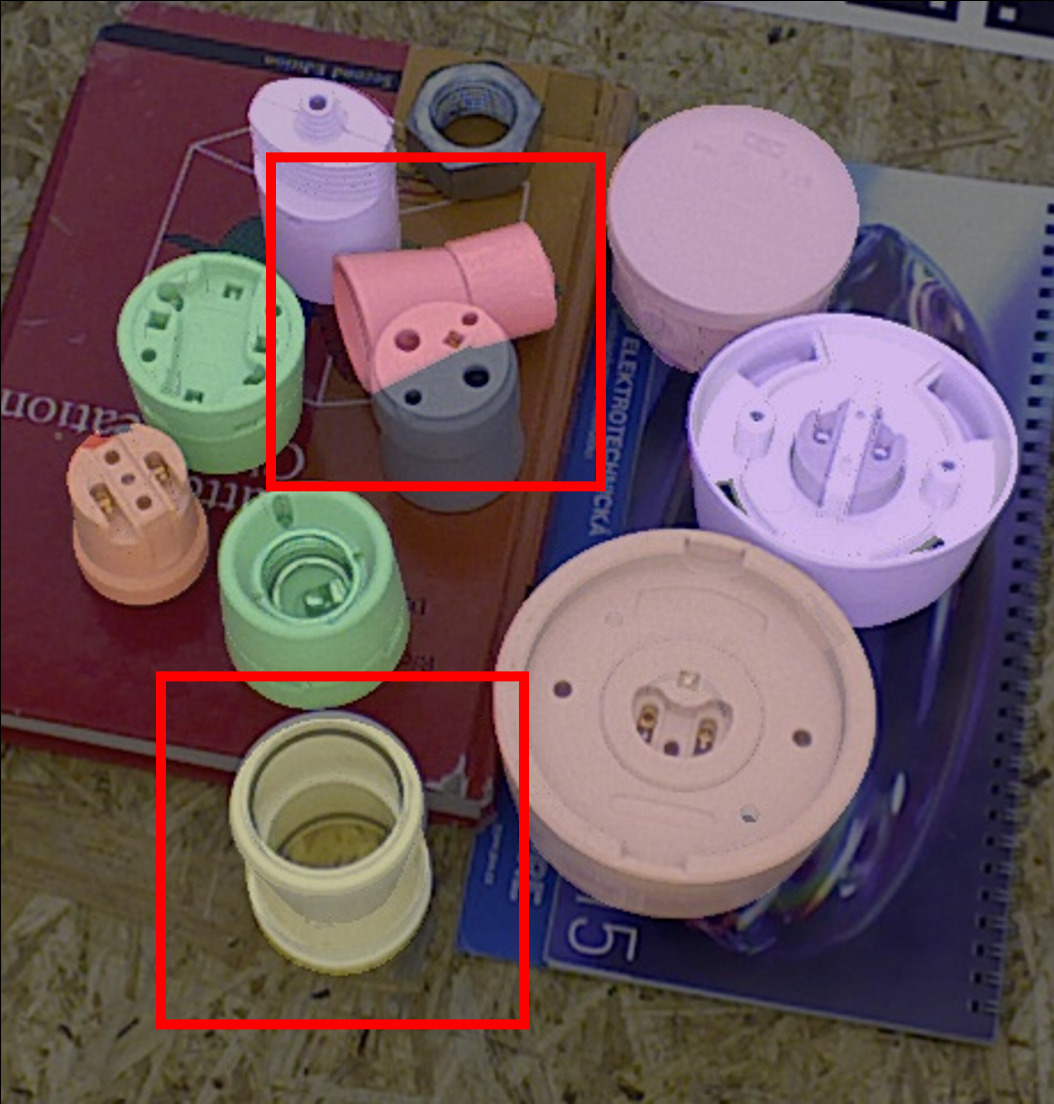}
  \caption{\label{subfig: overlayb}}
\end{subfigure}
\begin{subfigure}[b]{0.16\textwidth}
  \includegraphics[width=\columnwidth]{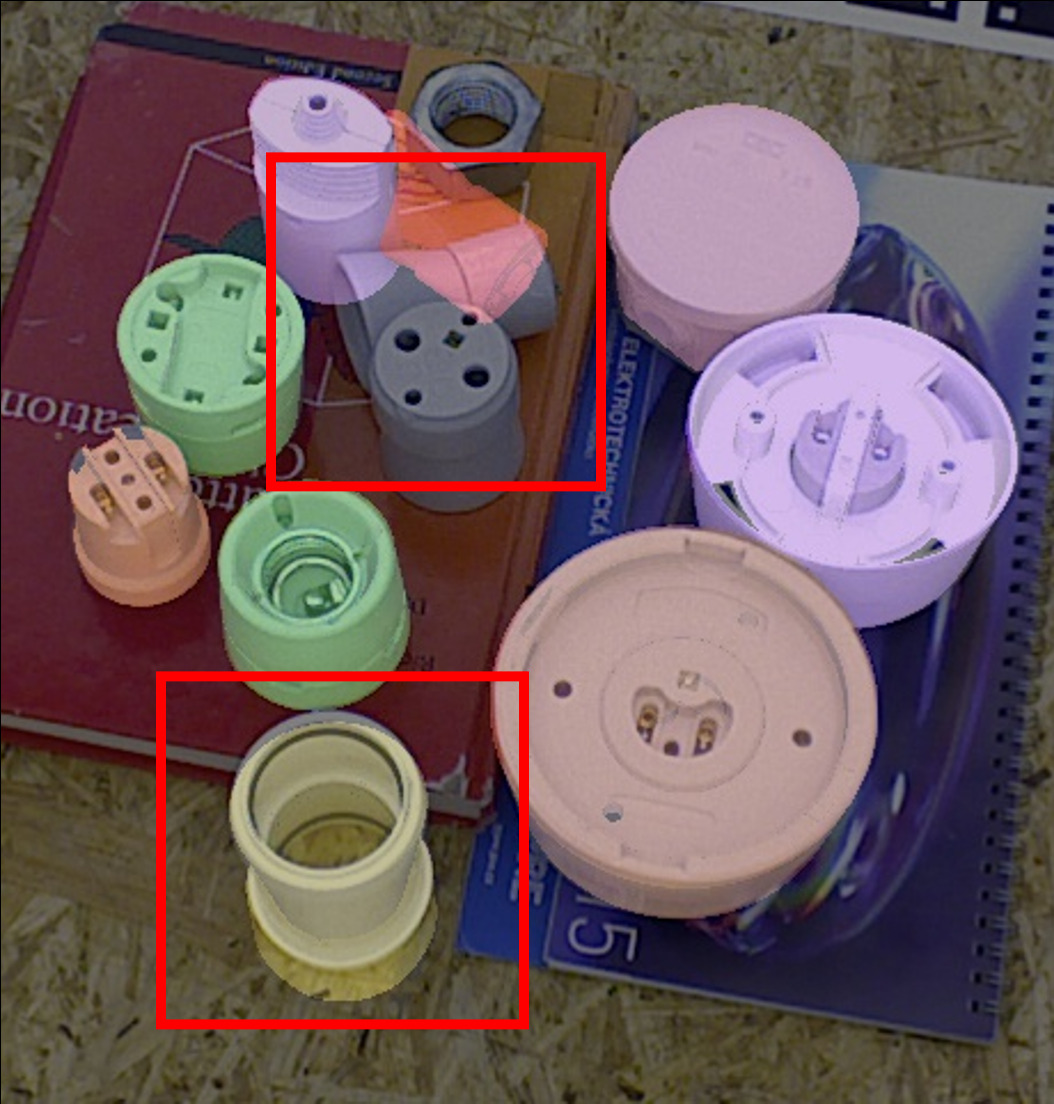}
  \caption{\label{subfig: overlayc}}
\end{subfigure}
\begin{subfigure}[b]{0.16\textwidth}
  \includegraphics[width=\columnwidth]{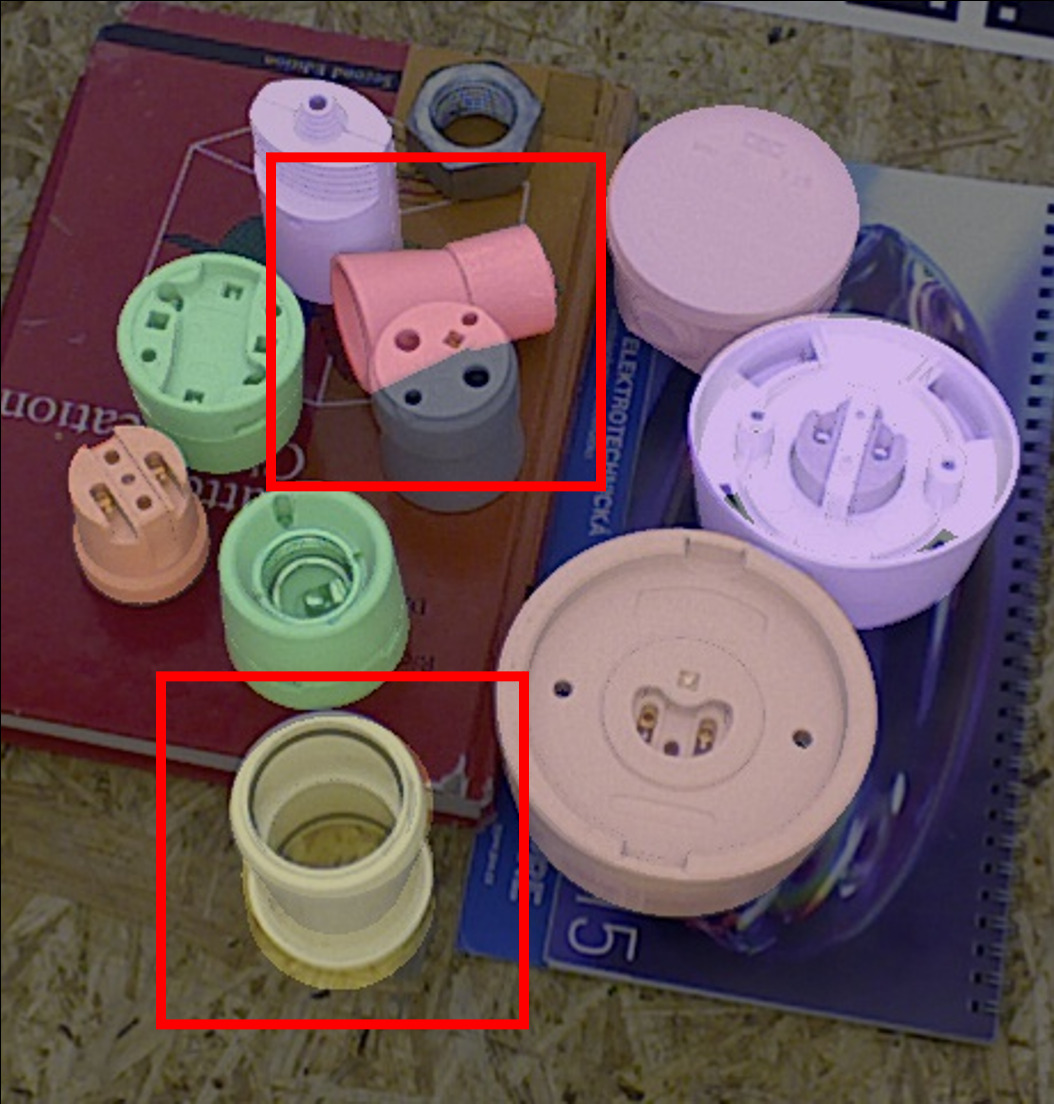}
  \caption{\label{subfig: overlayd}}
\end{subfigure}
\begin{subfigure}[b]{0.16\textwidth}
  \includegraphics[width=\columnwidth]{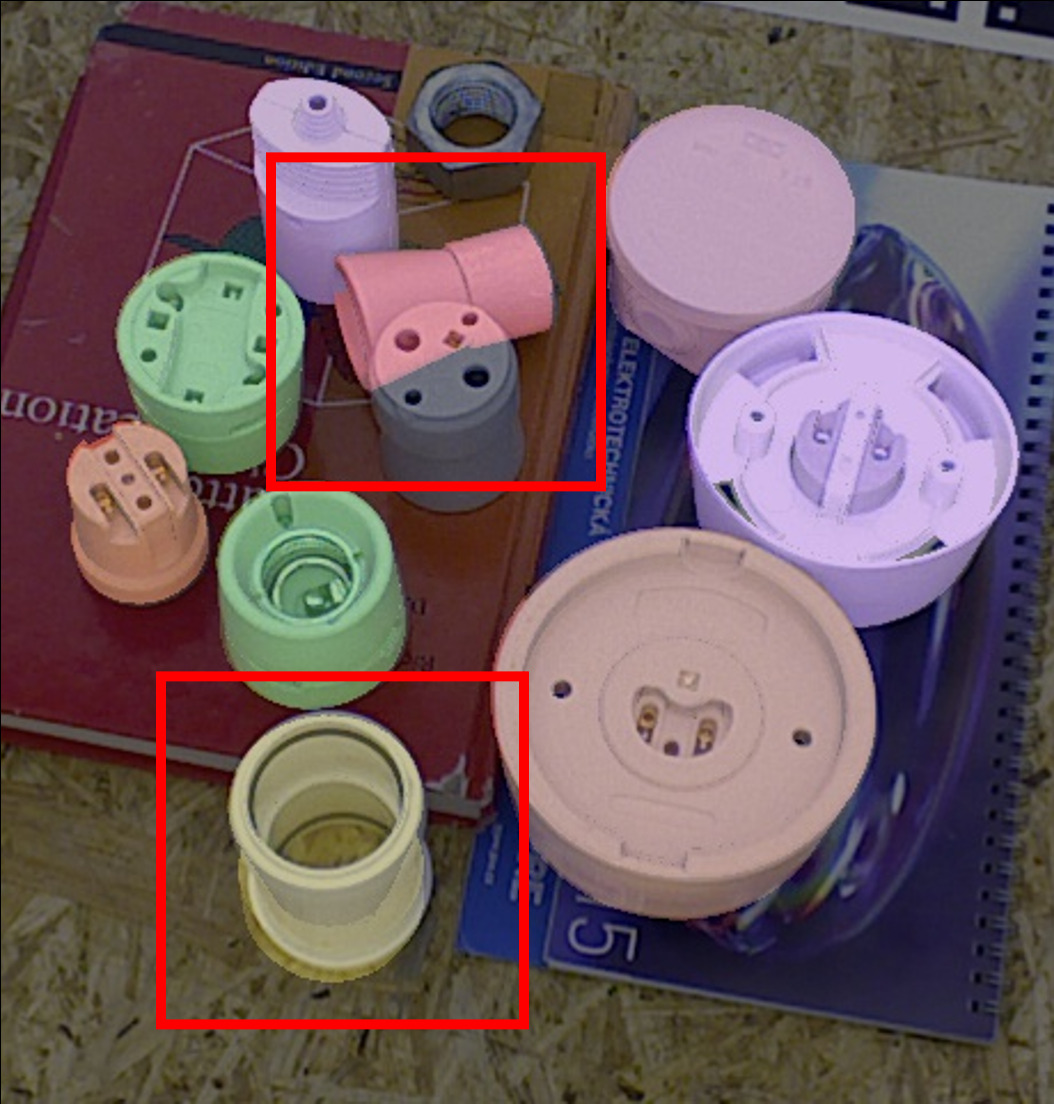}
  \caption{\label{subfig: overlaye}}
\end{subfigure}
\caption{Qualitative comparison of results on T-LESS: (a) Original RGB image, (b) MRC-Net, (c) CosyPose initialized PFA~\cite{hu2022perspective}, (d) SC6D~\cite{cai2022sc6d}, and (e) ZebraPose~\cite{su2022zebrapose}. The object's 3D model is projected with estimated 6D pose and overlaid on original images with distinct colors. Red boxes denote cases where pose predictions are distinctly different across the methods. MRC-Net outperforms the state-of-art models particularly under heavy occlusion. (Best viewed when zoomed in.) }
\label{fig: meshoverlay}
\end{figure*}

\textbf{Qualitative comparison}. Figure~\ref{fig: meshoverlay} depicts the qualitative performance of our model on T-LESS test images, compared with three representative state-of-the-art models: (a) PFA initialized with Cosypose (a 2D-to-2D correspondence model)~\cite{hu2022perspective}, (b) SC6D (a parallel classification and regression model)~\cite{cai2022sc6d}, and (c) ZebraPose (a 2D-to-3D indirect method)~\cite{su2022zebrapose}. For each object in the scene, we render the corresponding CAD model using the predicted pose to visualize its accuracy.  Our method is capable of accurately predicting object poses even under heavy occlusions, and is able to discern different objects or multiple instances of the same object even when they are nearby. In contrast, other methods predict inaccurate poses under these challenges. More visual results and failure cases are included in the supplementary material.

\textbf{Runtime analysis.} Table~\ref{tab:run_time} compares run time with recent methods for single object pose estimation from a $256\times 256$ image crop. 
As expected, our method is significantly faster than iterative refinement methods, even without accounting for their initialization step. MRC-Net is slower than SC6D and GDR-Net due to its on-the-fly rendering and two stage inference; however it significantly outperforms these two methods on accuracy, and can be considered viable for real-time operation at 16 frames-per-second.

\begin{table}
\setlength{\tabcolsep}{4.6pt}
\centering
\begin{tabularx}{\linewidth}{c|cccccc}
\toprule
Methods & \cite{lipson2022coupled}$^\ast$ & \cite{hu2022perspective}$^\ast$ & \cite{SurfEmb} & \cite{wang2021gdr} & \cite{cai2022sc6d} & MRC-Net \\
\midrule
Time (ms) & 2542 & 88 & 1121 & 26 & 25 & 61\\
\bottomrule
\end{tabularx}
\caption{\textbf{Runtime comparisons.} We run CIR~\cite{lipson2022coupled}, PFA~\cite{hu2022perspective}, SurfEmb~\cite{SurfEmb}, GDR-Net~\cite{wang2021gdr}, SC6D~\cite{cai2022sc6d} and MRC-Net (Ours) on the same AWS EC2 P3.2Xlarge instance, and report the average runtime in milliseconds to infer one object pose. For methods with $^\ast$, the time required for pose initialization, typically involving the execution of an additional off-the-shelf network, is not included.}\label{tab:run_time}
\end{table}

\section{Conclusions and Limitations}\label{sec:conclusions}
We introduced a novel architecture for single-shot 6-DoF pose estimation from single view RGB images, wherein pose classification acts as a conditioning input to residual pose regression. A critical part of the conditioning is a multi-scale correlation layer that captures and transfers residual image features from classification to regression. Extensive experiments on four challenging datasets demonstrate the superior performance of our method. As with other techniques~\cite{li2018deepim,SurfEmb}, our method suffers from degraded performance when the CAD models are inaccurate as it relies on the render-and-compare technique. Potential solutions involve enforcing stronger augmentations or developing model-agnostic techniques. Although our focus in this work is exclusively on RGB inputs, extending the framework to incorporate depth maps is an interesting future research direction to pursue.

\newpage
{\small
\bibliographystyle{ieeenat_fullname}
\bibliography{egbib}

\begin{thebibliography}{57}
\providecommand{\natexlab}[1]{#1}
\providecommand{\url}[1]{\texttt{#1}}
\expandafter\ifx\csname urlstyle\endcsname\relax
  \providecommand{\doi}[1]{doi: #1}\else
  \providecommand{\doi}{doi: \begingroup \urlstyle{rm}\Url}\fi

\bibitem[Brachmann et~al.(2014)Brachmann, Krull, Michel, Gumhold, Shotton, and Rother]{brachmann2014learning}
Eric Brachmann, Alexander Krull, Frank Michel, Stefan Gumhold, Jamie Shotton, and Carsten Rother.
\newblock Learning 6d object pose estimation using 3d object coordinates.
\newblock In \emph{European Conference on Computer Vision}, pages 536--551. Springer, 2014.

\bibitem[Cai et~al.(2022)Cai, Heikkil{\"a}, and Rahtu]{cai2022sc6d}
Dingding Cai, Janne Heikkil{\"a}, and Esa Rahtu.
\newblock {SC6D}: Symmetry-agnostic and correspondence-free 6d object pose estimation.
\newblock In \emph{International Conference on 3D Vision}, pages 536--546. IEEE, 2022.

\bibitem[Castro and Kim(2023)]{castro2023crt}
Pedro Castro and Tae-Kyun Kim.
\newblock {CRT-6D}: Fast 6d object pose estimation with cascaded refinement transformers.
\newblock In \emph{Proceedings of the IEEE Winter Conference on Applications of Computer Vision}, pages 5746--5755, 2023.

\bibitem[Chen et~al.(2022)Chen, Wang, Wang, Tian, Xiong, and Li]{chen2022epro}
Hansheng Chen, Pichao Wang, Fan Wang, Wei Tian, Lu Xiong, and Hao Li.
\newblock {EP}ro-{P}n{P}: Generalized end-to-end probabilistic perspective-n-points for monocular object pose estimation.
\newblock In \emph{Proceedings of the IEEE Conference on Computer Vision and Pattern Recognition}, pages 2781--2790, 2022.

\bibitem[Chen et~al.(2017)Chen, Papandreou, Kokkinos, Murphy, and Yuille]{chen2017deeplab}
Liang-Chieh Chen, George Papandreou, Iasonas Kokkinos, Kevin Murphy, and Alan~L Yuille.
\newblock Deep{L}ab: Semantic image segmentation with deep convolutional nets, atrous convolution, and fully connected crfs.
\newblock \emph{IEEE Transactions on Pattern Analysis and Machine Intelligence}, 40\penalty0 (4):\penalty0 834--848, 2017.

\bibitem[Collet and Srinivasa(2010)]{collet2010efficient}
Alvaro Collet and Siddhartha~S Srinivasa.
\newblock Efficient multi-view object recognition and full pose estimation.
\newblock In \emph{IEEE International Conference on Robotics and Automation}, pages 2050--2055. IEEE, 2010.

\bibitem[Di et~al.(2021)Di, Manhardt, Wang, Ji, Navab, and Tombari]{di2021so}
Yan Di, Fabian Manhardt, Gu Wang, Xiangyang Ji, Nassir Navab, and Federico Tombari.
\newblock {SO-Pose}: Exploiting self-occlusion for direct 6d pose estimation.
\newblock In \emph{Proceedings of the IEEE International Conference on Computer Vision}, pages 12396--12405, 2021.

\bibitem[Dosovitskiy et~al.(2015)Dosovitskiy, Fischer, Ilg, Hausser, Hazirbas, Golkov, Van Der~Smagt, Cremers, and Brox]{dosovitskiy2015flownet}
Alexey Dosovitskiy, Philipp Fischer, Eddy Ilg, Philip Hausser, Caner Hazirbas, Vladimir Golkov, Patrick Van Der~Smagt, Daniel Cremers, and Thomas Brox.
\newblock Flownet: Learning optical flow with convolutional networks.
\newblock In \emph{Proceedings of the IEEE International Conference on Computer Vision}, pages 2758--2766, 2015.

\bibitem[Drost et~al.(2017)Drost, Ulrich, Bergmann, Hartinger, and Steger]{drost2017introducing}
Bertram Drost, Markus Ulrich, Paul Bergmann, Philipp Hartinger, and Carsten Steger.
\newblock Introducing {MVT}ec {ITODD}-a dataset for 3d object recognition in industry.
\newblock In \emph{Proceedings of the IEEE International Conference on Computer Vision Workshops}, pages 2200--2208, 2017.

\bibitem[Guo et~al.(2023)Guo, Hu, Alvarez, and Salzmann]{guo2023knowledge}
Shuxuan Guo, Yinlin Hu, Jose~M. Alvarez, and Mathieu Salzmann.
\newblock Knowledge distillation for 6d pose estimation by aligning distributions of local predictions.
\newblock In \emph{Proceedings of the IEEE Conference on Computer Vision and Pattern Recognition}, pages 18633--18642, 2023.

\bibitem[Hai et~al.(2023{\natexlab{a}})Hai, Song, Li, and Hu]{hai2023shape}
Yang Hai, Rui Song, Jiaojiao Li, and Yinlin Hu.
\newblock Shape-constraint recurrent flow for 6d object pose estimation.
\newblock In \emph{Proceedings of the IEEE Conference on Computer Vision and Pattern Recognition}, pages 4831--4840, 2023{\natexlab{a}}.

\bibitem[Hai et~al.(2023{\natexlab{b}})Hai, Song, Li, Salzmann, and Hu]{hai2023rigidity}
Yang Hai, Rui Song, Jiaojiao Li, Mathieu Salzmann, and Yinlin Hu.
\newblock Rigidity-aware detection for 6d object pose estimation.
\newblock In \emph{Proceedings of the IEEE Conference on Computer Vision and Pattern Recognition}, pages 8927--8936, 2023{\natexlab{b}}.

\bibitem[Haugaard and Buch(2022)]{SurfEmb}
Rasmus~Laurvig Haugaard and Anders~Glent Buch.
\newblock Surfemb: Dense and continuous correspondence distributions for object pose estimation with learnt surface embeddings.
\newblock In \emph{Proceedings of the IEEE Conference on Computer Vision and Pattern Recognition}, pages 6749--6758, 2022.

\bibitem[He et~al.(2017)He, Gkioxari, Doll{\'a}r, and Girshick]{he2017mask}
Kaiming He, Georgia Gkioxari, Piotr Doll{\'a}r, and Ross Girshick.
\newblock Mask {R-CNN}.
\newblock In \emph{Proceedings of the IEEE International Conference on Computer Vision}, pages 2961--2969, 2017.

\bibitem[Hinterstoisser et~al.(2011)Hinterstoisser, Holzer, Cagniart, Ilic, Konolige, Navab, and Lepetit]{hinterstoisser2011multimodal}
Stefan Hinterstoisser, Stefan Holzer, Cedric Cagniart, Slobodan Ilic, Kurt Konolige, Nassir Navab, and Vincent Lepetit.
\newblock Multimodal templates for real-time detection of texture-less objects in heavily cluttered scenes.
\newblock In \emph{Proceedings of the IEEE International Conference on Computer Vision}, pages 858--865. IEEE, 2011.

\bibitem[Hodan et~al.(2017)Hodan, Haluza, Obdr{\v{z}}{\'a}lek, Matas, Lourakis, and Zabulis]{hodan2017tless}
Tom{\'a}{\v{s}} Hodan, Pavel Haluza, {\v{S}}tep{\'a}n Obdr{\v{z}}{\'a}lek, Jiri Matas, Manolis Lourakis, and Xenophon Zabulis.
\newblock T-{LESS}: An {RGB-D} dataset for 6d pose estimation of texture-less objects.
\newblock In \emph{IEEE Winter Conference on Applications of Computer Vision}, pages 880--888. IEEE, 2017.

\bibitem[Hodan et~al.(2020)Hodan, Barath, and Matas]{hodan2020epos}
Tomas Hodan, Daniel Barath, and Jiri Matas.
\newblock {EPOS}: Estimating 6d pose of objects with symmetries.
\newblock In \emph{Proceedings of the IEEE Conference on Computer Vision and Pattern Recognition}, pages 11703--11712, 2020.

\bibitem[Hoda{\v{n}} et~al.(2020)Hoda{\v{n}}, Sundermeyer, Drost, Labb{\'e}, Brachmann, Michel, Rother, and Matas]{hodavn2020bop}
Tom{\'a}{\v{s}} Hoda{\v{n}}, Martin Sundermeyer, Bertram Drost, Yann Labb{\'e}, Eric Brachmann, Frank Michel, Carsten Rother, and Ji{\v{r}}{\'\i} Matas.
\newblock {BOP} challenge 2020 on 6d object localization.
\newblock In \emph{European Conference on Computer Vision}, pages 577--594. Springer, 2020.

\bibitem[Hu et~al.(2019)Hu, Hugonot, Fua, and Salzmann]{hu2019segmentation}
Yinlin Hu, Joachim Hugonot, Pascal Fua, and Mathieu Salzmann.
\newblock Segmentation-driven 6d object pose estimation.
\newblock In \emph{Proceedings of the IEEE Conference on Computer Vision and Pattern Recognition}, pages 3385--3394, 2019.

\bibitem[Hu et~al.(2020)Hu, Fua, Wang, and Salzmann]{hu2020single}
Yinlin Hu, Pascal Fua, Wei Wang, and Mathieu Salzmann.
\newblock Single-stage 6d object pose estimation.
\newblock In \emph{Proceedings of the IEEE Conference on Computer Vision and Pattern Recognition}, pages 2930--2939, 2020.

\bibitem[Hu et~al.(2021)Hu, Speierer, Jakob, Fua, and Salzmann]{hu2021wide}
Yinlin Hu, Sebastien Speierer, Wenzel Jakob, Pascal Fua, and Mathieu Salzmann.
\newblock Wide-depth-range 6d object pose estimation in space.
\newblock In \emph{Proceedings of the IEEE Conference on Computer Vision and Pattern Recognition}, pages 15870--15879, 2021.

\bibitem[Hu et~al.(2022)Hu, Fua, and Salzmann]{hu2022perspective}
Yinlin Hu, Pascal Fua, and Mathieu Salzmann.
\newblock Perspective flow aggregation for data-limited 6d object pose estimation.
\newblock In \emph{European Conference on Computer Vision}, pages 89--106. Springer, 2022.

\bibitem[Huang et~al.(2022)Huang, Hodan, Ma, Zhang, Tran, Twigg, Wu, Yuan, Keskin, and Wang]{huang2022neural}
Lin Huang, Tomas Hodan, Lingni Ma, Linguang Zhang, Luan Tran, Christopher Twigg, Po-Chen Wu, Junsong Yuan, Cem Keskin, and Robert Wang.
\newblock Neural correspondence field for object pose estimation.
\newblock In \emph{European Conference on Computer Vision}, pages 585--603. Springer, 2022.

\bibitem[Iwase et~al.(2021)Iwase, Liu, Khirodkar, Yokota, and Kitani]{iwase2021repose}
Shun Iwase, Xingyu Liu, Rawal Khirodkar, Rio Yokota, and Kris~M Kitani.
\newblock Re{POSE}: Fast 6d object pose refinement via deep texture rendering.
\newblock In \emph{Proceedings of the IEEE International Conference on Computer Vision}, pages 3303--3312, 2021.

\bibitem[Kehl et~al.(2017)Kehl, Manhardt, Tombari, Ilic, and Navab]{kehl2017ssd}
Wadim Kehl, Fabian Manhardt, Federico Tombari, Slobodan Ilic, and Nassir Navab.
\newblock {SSD-6D}: Making rgb-based 3d detection and 6d pose estimation great again.
\newblock In \emph{Proceedings of the IEEE International Conference on Computer Vision}, pages 1521--1529, 2017.

\bibitem[Labb{\'e} et~al.(2020)Labb{\'e}, Carpentier, Aubry, and Sivic]{labb2020cosypose}
Yann Labb{\'e}, Justin Carpentier, Mathieu Aubry, and Josef Sivic.
\newblock Cosy{P}ose: Consistent multi-view multi-object 6d pose estimation.
\newblock In \emph{European Conference on Computer Vision}, pages 574--591. Springer, 2020.

\bibitem[Labb\'e et~al.(2022)Labb\'e, Manuelli, Mousavian, Tyree, Birchfield, Tremblay, Carpentier, Aubry, Fox, and Sivic]{labbe2022megapose}
Yann Labb\'e, Lucas Manuelli, Arsalan Mousavian, Stephen Tyree, Stan Birchfield, Jonathan Tremblay, Justin Carpentier, Mathieu Aubry, Dieter Fox, and Josef Sivic.
\newblock {MegaPose}: 6d pose estimation of novel objects via render \& compare.
\newblock In \emph{Proceedings of the 6th Conference on Robot Learning}, 2022.

\bibitem[Li et~al.(2018)Li, Wang, Ji, Xiang, and Fox]{li2018deepim}
Yi Li, Gu Wang, Xiangyang Ji, Yu Xiang, and Dieter Fox.
\newblock {DeepIM}: Deep iterative matching for 6d pose estimation.
\newblock In \emph{Proceedings of the European Conference on Computer Vision}, pages 683--698, 2018.

\bibitem[Li et~al.(2019)Li, Wang, and Ji]{li2019cdpn}
Zhigang Li, Gu Wang, and Xiangyang Ji.
\newblock {CDPN:} coordinates-based disentangled pose network for real-time rgb-based 6-{DoF} object pose estimation.
\newblock In \emph{IEEE International Conference on Computer Vision}, pages 7677--7686, 2019.

\bibitem[Li et~al.(2022)Li, Liu, Zhang, Xu, and Yan]{li2022cliff}
Zhihao Li, Jianzhuang Liu, Zhensong Zhang, Songcen Xu, and Youliang Yan.
\newblock {CLIFF}: Carrying location information in full frames into human pose and shape estimation.
\newblock In \emph{European Conference on Computer Vision}, pages 590--606. Springer, 2022.

\bibitem[Lian and Kim(2023)]{lian2023checkerpose}
Ruyi Lian and Tae-Kyun Kim.
\newblock {CheckerPose}: Progressive dense keypoint localization for object pose estimation with graph neural network.
\newblock In \emph{Proceedings of the IEEE International Conference on Computer Vision}, pages 14022--14033, 2023.

\bibitem[Lin et~al.(2017)Lin, Goyal, Girshick, He, and Doll{\'a}r]{lin2017focal}
Tsung-Yi Lin, Priya Goyal, Ross Girshick, Kaiming He, and Piotr Doll{\'a}r.
\newblock Focal loss for dense object detection.
\newblock In \emph{Proceedings of the IEEE International Conference on Computer Vision}, pages 2980--2988, 2017.

\bibitem[Lipson et~al.(2022)Lipson, Teed, Goyal, and Deng]{lipson2022coupled}
Lahav Lipson, Zachary Teed, Ankit Goyal, and Jia Deng.
\newblock Coupled iterative refinement for 6d multi-object pose estimation.
\newblock In \emph{Proceedings of the IEEE Conference on Computer Vision and Pattern Recognition}, pages 6728--6737, 2022.

\bibitem[Liu et~al.(2023)Liu, Hu, and Salzmann]{liu2023linear}
Fulin Liu, Yinlin Hu, and Mathieu Salzmann.
\newblock Linear-covariance loss for end-to-end learning of 6d pose estimation.
\newblock In \emph{Proceedings of the IEEE International Conference on Computer Vision}, pages 14107--14117, 2023.

\bibitem[Loshchilov and Hutter(2019)]{loshchilov2018decoupled}
Ilya Loshchilov and Frank Hutter.
\newblock Decoupled weight decay regularization.
\newblock In \emph{International Conference on Learning Representations}, 2019.

\bibitem[Manhardt et~al.(2018)Manhardt, Kehl, Navab, and Tombari]{manhardt2018deep}
Fabian Manhardt, Wadim Kehl, Nassir Navab, and Federico Tombari.
\newblock Deep model-based 6d pose refinement in rgb.
\newblock In \emph{Proceedings of the European Conference on Computer Vision}, pages 800--815, 2018.

\bibitem[Mousavian et~al.(2017)Mousavian, Anguelov, Flynn, and Kosecka]{mousavian20173d}
Arsalan Mousavian, Dragomir Anguelov, John Flynn, and Jana Kosecka.
\newblock {3D} bounding box estimation using deep learning and geometry.
\newblock In \emph{Proceedings of the IEEE Conference on computer Vision and Pattern Recognition}, pages 7074--7082, 2017.

\bibitem[Nocedal and Wright(1999)]{nocedal1999numerical}
Jorge Nocedal and Stephen~J Wright.
\newblock \emph{Numerical optimization}.
\newblock Springer, 1999.

\bibitem[Park and Cho(2022)]{park2022dprost}
Jaewoo Park and Nam~Ik Cho.
\newblock {DProST}: Dynamic projective spatial transformer network for 6d pose estimation.
\newblock In \emph{European Conference on Computer Vision}, pages 363--379. Springer, 2022.

\bibitem[Peng et~al.(2019)Peng, Liu, Huang, Zhou, and Bao]{peng2019pvnet}
Sida Peng, Yuan Liu, Qixing Huang, Xiaowei Zhou, and Hujun Bao.
\newblock {PVNet}: Pixel-wise voting network for 6dof pose estimation.
\newblock In \emph{Proceedings of the IEEE Conference on Computer Vision and Pattern Recognition}, pages 4561--4570, 2019.

\bibitem[Rad and Lepetit(2017)]{rad2017bb8}
Mahdi Rad and Vincent Lepetit.
\newblock {BB8}: A scalable, accurate, robust to partial occlusion method for predicting the 3d poses of challenging objects without using depth.
\newblock In \emph{Proceedings of the IEEE International Conference on Computer Vision}, pages 3828--3836, 2017.

\bibitem[Ravi et~al.(2020)Ravi, Reizenstein, Novotny, Gordon, Lo, Johnson, and Gkioxari]{ravi2020pytorch3d}
Nikhila Ravi, Jeremy Reizenstein, David Novotny, Taylor Gordon, Wan-Yen Lo, Justin Johnson, and Georgia Gkioxari.
\newblock Accelerating 3d deep learning with pytorch3d.
\newblock \emph{arXiv:2007.08501}, 2020.

\bibitem[Risholm et~al.(2021)Risholm, Ivarsen, Haugholt, and Mohammed]{risholm2021underwater}
Petter Risholm, Peter~{\O}rnulf Ivarsen, Karl~Henrik Haugholt, and Ahmed Mohammed.
\newblock Underwater marker-based pose-estimation with associated uncertainty.
\newblock In \emph{Proceedings of the IEEE International Conference on Computer Vision}, pages 3713--3721, 2021.

\bibitem[Shugurov et~al.(2021)Shugurov, Zakharov, and Ilic]{shugurov2021dpodv2}
Ivan Shugurov, Sergey Zakharov, and Slobodan Ilic.
\newblock {DPOD}v2: Dense correspondence-based 6 dof pose estimation.
\newblock \emph{IEEE Transactions on Pattern Analysis and Machine Intelligence}, 2021.

\bibitem[Simonelli et~al.(2019)Simonelli, Bulo, Porzi, L{\'o}pez-Antequera, and Kontschieder]{simonelli2019disentangling}
Andrea Simonelli, Samuel~Rota Bulo, Lorenzo Porzi, Manuel L{\'o}pez-Antequera, and Peter Kontschieder.
\newblock Disentangling monocular 3d object detection.
\newblock In \emph{Proceedings of the IEEE International Conference on Computer Vision}, pages 1991--1999, 2019.

\bibitem[Song et~al.(2020)Song, Song, and Huang]{song2020hybridpose}
Chen Song, Jiaru Song, and Qixing Huang.
\newblock Hybrid{P}ose: 6d object pose estimation under hybrid representations.
\newblock In \emph{Proceedings of the IEEE Conference on Computer Vision and Pattern Recognition}, pages 431--440, 2020.

\bibitem[Su et~al.(2022)Su, Saleh, Fetzer, Rambach, Navab, Busam, Stricker, and Tombari]{su2022zebrapose}
Yongzhi Su, Mahdi Saleh, Torben Fetzer, Jason Rambach, Nassir Navab, Benjamin Busam, Didier Stricker, and Federico Tombari.
\newblock Zebra{P}ose: Coarse to fine surface encoding for 6dof object pose estimation.
\newblock In \emph{Proceedings of the IEEE Conference on Computer Vision and Pattern Recognition}, pages 6738--6748, 2022.

\bibitem[Sun et~al.(2018)Sun, Yang, Liu, and Kautz]{sun2018pwc}
Deqing Sun, Xiaodong Yang, Ming-Yu Liu, and Jan Kautz.
\newblock {PWC-Net}: Cnns for optical flow using pyramid, warping, and cost volume.
\newblock In \emph{Proceedings of the IEEE Conference on Computer Vision and Pattern Recognition}, pages 8934--8943, 2018.

\bibitem[Sundermeyer et~al.(2018)Sundermeyer, Marton, Durner, Brucker, and Triebel]{sundermeyer2018implicit}
Martin Sundermeyer, Zoltan-Csaba Marton, Maximilian Durner, Manuel Brucker, and Rudolph Triebel.
\newblock Implicit 3d orientation learning for 6d object detection from rgb images.
\newblock In \emph{European Conference on Computer Vision}, pages 699--715, 2018.

\bibitem[Sundermeyer et~al.(2020)Sundermeyer, Durner, Puang, Marton, Vaskevicius, Arras, and Triebel]{sundermeyer2020multi}
Martin Sundermeyer, Maximilian Durner, En~Yen Puang, Zoltan-Csaba Marton, Narunas Vaskevicius, Kai~O Arras, and Rudolph Triebel.
\newblock Multi-path learning for object pose estimation across domains.
\newblock In \emph{Proceedings of the IEEE Conference on Computer Vision and Pattern Recognition}, pages 13916--13925, 2020.

\bibitem[Teed and Deng(2020)]{teed2020raft}
Zachary Teed and Jia Deng.
\newblock {RAFT}: Recurrent all-pairs field transforms for optical flow.
\newblock In \emph{European Conference on Computer Vision}, pages 402--419. Springer, 2020.

\bibitem[Wang et~al.(2021)Wang, Manhardt, Tombari, and Ji]{wang2021gdr}
Gu Wang, Fabian Manhardt, Federico Tombari, and Xiangyang Ji.
\newblock {GDR-Net}: Geometry-guided direct regression network for monocular 6d object pose estimation.
\newblock In \emph{Proceedings of the IEEE Conference on Computer Vision and Pattern Recognition}, pages 16611--16621, 2021.

\bibitem[Xiang et~al.(2018)Xiang, Schmidt, Narayanan, and Fox]{xiang2017posecnn}
Yu Xiang, Tanner Schmidt, Venkatraman Narayanan, and Dieter Fox.
\newblock {PoseCNN}: A convolutional neural network for 6d object pose estimation in cluttered scenes.
\newblock \emph{Robotics: Science and Systems}, 2018.

\bibitem[Xu et~al.(2022)Xu, Lin, Zhang, Wang, and Li]{xu2022rnnpose}
Yan Xu, Kwan-Yee Lin, Guofeng Zhang, Xiaogang Wang, and Hongsheng Li.
\newblock {RNNPose}: Recurrent 6-dof object pose refinement with robust correspondence field estimation and pose optimization.
\newblock In \emph{Proceedings of the IEEE Conference on Computer Vision and Pattern Recognition}, pages 14880--14890, 2022.

\bibitem[Yershova et~al.(2010)Yershova, Jain, Lavalle, and Mitchell]{yershova2010generating}
Anna Yershova, Swati Jain, Steven~M Lavalle, and Julie~C Mitchell.
\newblock Generating uniform incremental grids on {SO(3)} using the {H}opf fibration.
\newblock \emph{The International Journal of Robotics Research}, 29\penalty0 (7):\penalty0 801--812, 2010.

\bibitem[Zhao et~al.(2023)Zhao, Wei, Shi, Tan, Li, Ren, Wei, Yang, and Pu]{zhao2023learning}
Heng Zhao, Shenxing Wei, Dahu Shi, Wenming Tan, Zheyang Li, Ye Ren, Xing Wei, Yi Yang, and Shiliang Pu.
\newblock Learning symmetry-aware geometry correspondences for 6d object pose estimation.
\newblock In \emph{Proceedings of the IEEE International Conference on Computer Vision}, pages 14045--14054, 2023.

\bibitem[Zhou et~al.(2019)Zhou, Barnes, Lu, Yang, and Li]{zhou2019continuity}
Yi Zhou, Connelly Barnes, Jingwan Lu, Jimei Yang, and Hao Li.
\newblock On the continuity of rotation representations in neural networks.
\newblock In \emph{Proceedings of the IEEE Conference on Computer Vision and Pattern Recognition}, pages 5745--5753, 2019.

\end{thebibliography}
}

\end{document}


\title{MRC-Net: 6-DoF Pose Estimation with MultiScale Residual Correlation\\ Supplementary Material}

\author{Yuelong Li \thanks{Equal contribution}\\
Amazon Inc.\\
{\tt\small yuell@amazon.com}
\and
Yafei Mao \footnotemark[1]\\
Amazon Inc.\\
{\tt\small yafeimao@amazon.com}
\and
Raja Bala\\
Amazon Inc.\\
{\tt\small rajabl@amazon.com}
\and
Sunil Hadap\\
Amazon Inc.\\
{\tt\small hadsunil@lab126.com}
}
\maketitle


\section{Implementation Details}\label{sec:details}
\subsection{Network Architecture}\label{ssec:architecture}
Referring to Fig. 2 in the main paper, we input a $256\times 256$ crop into a ResNet-34 backbone without pooling and fully-connected layers, yielding a $16\times 16\times 512$ feature tensor. The decoder conducts two rounds of $2\times$ upsampling on this tensor, incorporating low-level features via skip connections, resulting in a $64\times 64$ feature tensor. Following the approach of~\cite{sundermeyer2020multi, SurfEmb, cai2022sc6d}, we use individual decoders for each object, each producing a $64$-channel image feature and a binary segmentation mask indicating object visibility. These outputs are concatenated and downsampled through two consecutive convolution layers with a stride of two, forming a three-scale feature pyramid. This is processed through an ASPP block~\cite{chen2017deeplab} with dilation rates $r=2,3,4,6,8,12$, aligning with the rates in the ASPP block preceding the regression and classification heads, and capturing both local and global context. We do not track running means and variances across shared batch normalization layers, since real and rendered batches can have different statistics.

\begin{table}[h]
\centering
\setlength{\tabcolsep}{12pt}
  \begin{tabular}{cccc}
    \toprule
   \makecell{Input\\resolution} & \makecell{Input\\channels} & \makecell{Output\\channels} & Stride \\
    \midrule
    $64\times 64$ & $186$ & $128$ & $2$\\
    $32\times 32$ & $377$ & $256$ & $2$\\
    $16\times 16$ & $505$ & $256$ & $1$\\
    \bottomrule
  \end{tabular}
  \caption{Architecture details of the $3\times 3$ convolutions inside the MRC block. Each convolution layer is followed by group normalization~\cite{wu2018group} and the Swish activation function~\cite{ramachandran2018searching}.}\label{tab:mrc_details}
\end{table}

The feature pyramids from both real and rendered images serve as inputs to our novel MRC block. All $1\times 1$ convolutions within this block have both $128$ input and output channels. Additionally, a sequence of $3\times 3$ convolutions is employed to fuse real and correlation features. The architecture of these convolution layers is detailed in Table~\ref{tab:mrc_details}.

Finally, each individual task head comprises a simple two-layer multi-layer perceptron (MLP) with a hidden size of $256$.

\subsection{Rotation and Translation Representations}\label{ssec:partition}
To form a uniform partition of $SO(3)$, we follow the approach in~\cite{yershova2010generating} to generate $K=4608$ uniform grids in $SO(3)$. This involves using Hopf coordinates~\cite{yershova2010generating} to decompose 3D rotation into a 1D in-plane rotation (around the $z$ axis) and a 2D out-of-plane rotation. We then generate $m_2=192$ out-of-plane components using the formula from~\cite{gorski2005healpix} with $N_{\mathrm{side}}=4$, followed by $m_1=\sqrt{\pi m_2}\approx 24$ in-plane components.




For translation, we adopt the SITE format~\cite{li2019cdpn} to disambiguate the network prediction target based on local image patches. Our formulation of $\tau_z$ slightly differs from the original version:
\[
    \tau_z=\frac{t_z}{r\sqrt{f_x f_y}},
\]
where $f_x,f_y$ are the camera focal lengths and $r$ is the resizing ratio of the bounding box. This adjustment is made to normalize $\tau_z$ consistently across different cameras.

To measure distance between poses, we use the formulation in~\cite{labb2020cosypose} which computes vertex-based distances over CAD models. To ensure differentiability near the origin, we adopt smooth L1 loss~\cite{girshick2015fast} as a substitute of vertex distances. We also leverage the symmetry-aware formulation in~\cite{labb2020cosypose} to account for object symmetries.

To decouple relative rotation and translation between real and rendered images,we closely follow the approach of \cite{li2018deepim,labb2020cosypose},
differing only in the use of $\Delta\tau_z$ as an \emph{additive} residual for $\tau_z$ instead of a \emph{multiplicative} one. This modification has been found to enhance training stability.

\section{\tcre{Additional Ablation Studies}}\label{sec:ablations}

MRC-Net uses a Siamese structure for classification and regression branches. This enables real and rendered image features to be projected into a shared embedding space to accurately capture correspondences between them. To verify efficacy of this design choice, we conduct ablation experiments comparing Siamese and non-Siamese versions. The latter decouples the network weights in the classification and regression branches, as done in~\cite{labbe2022megapose}. We use similar strategies as~\cite{zhou2020tracking, labbe2022megapose} to train classifier and regressor separately. During training, the ground truth pose is perturbed synthetically by adding random noise to simulate classification errors. We then render the images based on these noisy poses. We follow the same noise schedule as described in~\cite{labbe2022megapose}. Comparisons across different methods are summarized in Table~\ref{tab:abl_siamese}.

The non-Siamese network with separate training exhibits a significant 11.4\% drop in average recall (AR) compared to the proposed Siamese model. Even with an end-to-end training strategy, it lags notably by 1.2\%. In contrast, the Siamese model achieves not only superior performance but also computational and storage efficiency by nearly halving parameters through parameter sharing.

\tcre{We next study the impact of the number of rotation buckets $K$. Our choice $K=4608 (N_{\mathrm{side}}=4)$ is the maximum value that we could fit into our GPU memory, which gives a relative angle of $14.7^{\circ}$ between adjacent rotation buckets~\cite{gorski2005healpix}. In our $SO(3)$ partitioning scheme (Section~\ref{ssec:partition}), alternative values for $K$ include $K=576 (N_{\mathrm{side}}=2)$ and $K=1944 (N_{\mathrm{side}}=3)$. Results are summarized in Table~\ref{tab:abl_K}. As $K$ increases, performance of our model improves slightly, with $K=4608$ achieving the highest AR scores. This validates our choice of $K$ and supports the assumption that finer classification can facilitate regression by providing a better pose initialization.}

\begin{table*}[h!]
    \begin{subtable}{\linewidth}
       \centering
        \begin{tabular}{c|ccc|c}
          \toprule
      Method & $\mathrm{AR_{VSD}}$ & $\mathrm{AR_{MSSD}}$ & $\mathrm{AR_{MSPD}}$ & $\mathrm{AR}$ \\
      \midrule
          Non-Siamese network with separate training & $56.1$ & $63.5$ & $77.4$ & $65.7$\\
          Non-Siamese network with end-to-end training & $68.9$ & $73.0$ & $85.8$ & $75.9$ \\
          Siamese network (Ours) & $70.6$ & $74.7$ & $86.0$ & $77.1$ \\
          \bottomrule
        \end{tabular}
        \caption{Ablation studies on Siamese and non-Siamese designs.}\label{tab:abl_siamese}
    \end{subtable}

    \begin{subtable}{\linewidth}
      \centering
      \begin{tabular}{c|ccc|c}
          \toprule
          $K$ & $\mathrm{AR_{VSD}}$ & $\mathrm{AR_{MSSD}}$ & $\mathrm{AR_{MSPD}}$ & $\mathrm{AR}$ \\
          \midrule
          $576$ & $70.3$ & $74.3$ & $85.6$ & $76.7$\\
          $1944$ & $70.4$ & $74.4$ & $85.8$ & $76.9$\\
          $4608$ & $70.6$ & $74.7$ & $86.0$ & $77.1$\\
          \bottomrule
      \end{tabular}
      \caption{Ablation studies on K, the number of rotation buckets.}\label{tab:abl_K}
    \end{subtable}
    \caption{\tcre{Additional ablation studies. Models are trained on T-LESS dataset using synthetic training set~\cite{hodan2017tless}.}}
\end{table*}

\section{Detailed Results on YCB-V}\label{sec:detailed_ycbv}
Table~\ref{tab:ycbv_per_obj} shows a comprehensive breakdown of per-object results for the AUC of ADD-S and AUC of ADD(-S). Previous works commonly use Mask RCNN detections provided by CosyPose~\cite{labb2020cosypose} or FCOS detections provided by CDPN~\cite{li2019cdpn} for their evaluations. Therefore, we present results for both. Compared to previous techniques, our method demonstrates notable improvements across most objects, resulting in an increase of +5.1\% in the average ADD-S AUC and +7.9\% in the average ADD(-S) AUC when combined with FCOS detections.

Further insights into the ADD(-S) metric for individual objects are presented in Table~\ref{tab:ycbv_adds}. MRC-Net outperforms prior models across various objects, leading to an +1.8\% improvement in average ADD(-S) recall when using FCOS detections.

\section{Additional Qualitative Evaluation}\label{sec:visual_examples}
\begin{figure*}[h!]
\centering
    \begin{subfigure}[b]{0.23\textwidth}
        \includegraphics[width=\columnwidth]{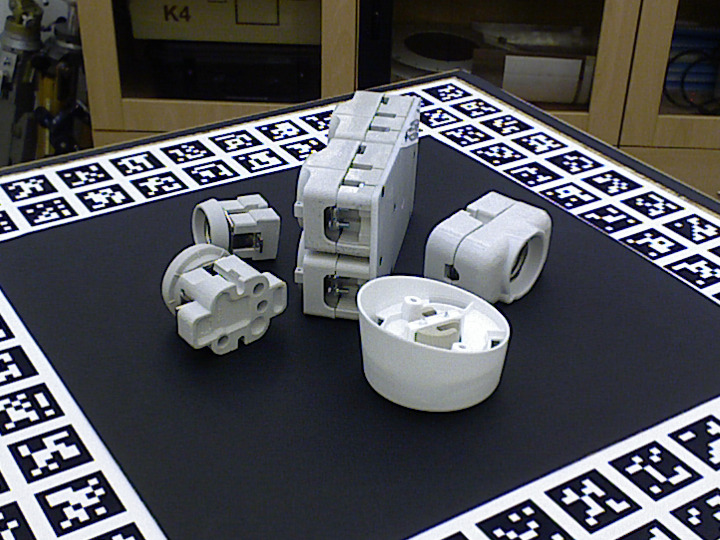}
        \end{subfigure}
    \begin{subfigure}[b]{0.23\textwidth}
        \includegraphics[width=\columnwidth]{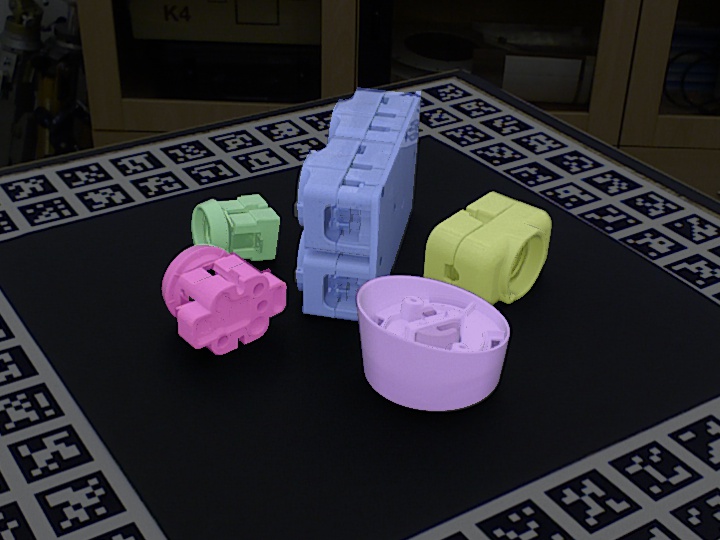}
        \end{subfigure}
    \begin{subfigure}[b]{0.23\textwidth}
        \includegraphics[width=\columnwidth]{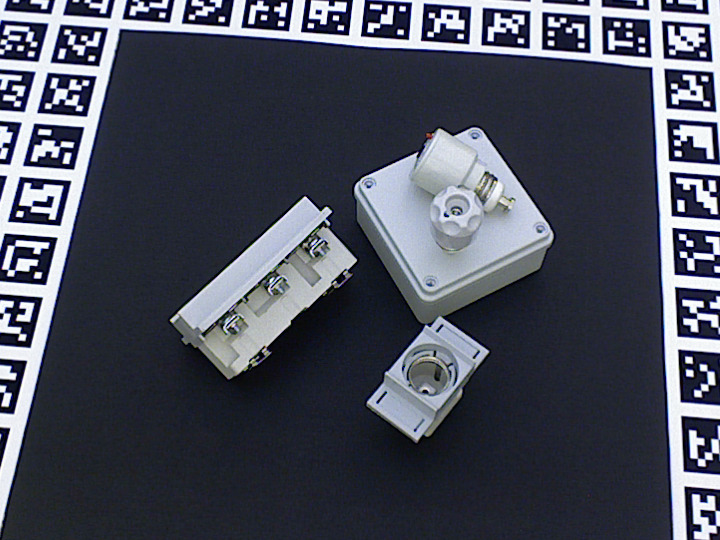}
        \end{subfigure}
    \begin{subfigure}[b]{0.23\textwidth}
        \includegraphics[width=\columnwidth]{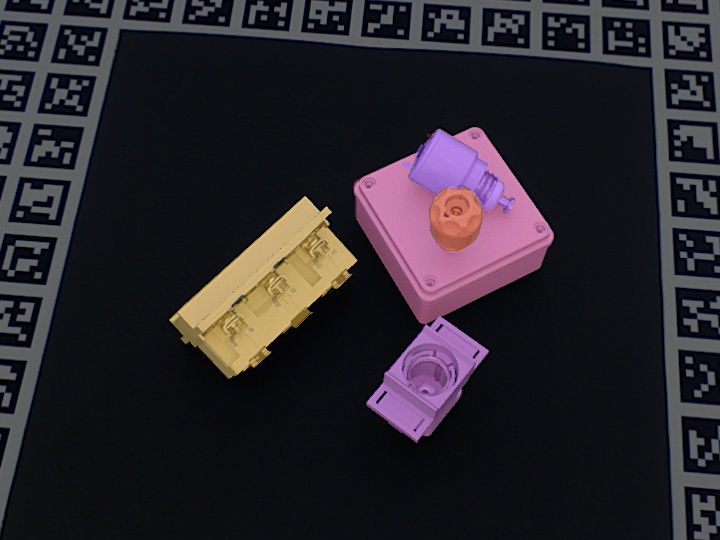}
        \end{subfigure}
    \begin{subfigure}[b]{0.23\textwidth}
        \includegraphics[width=\columnwidth]{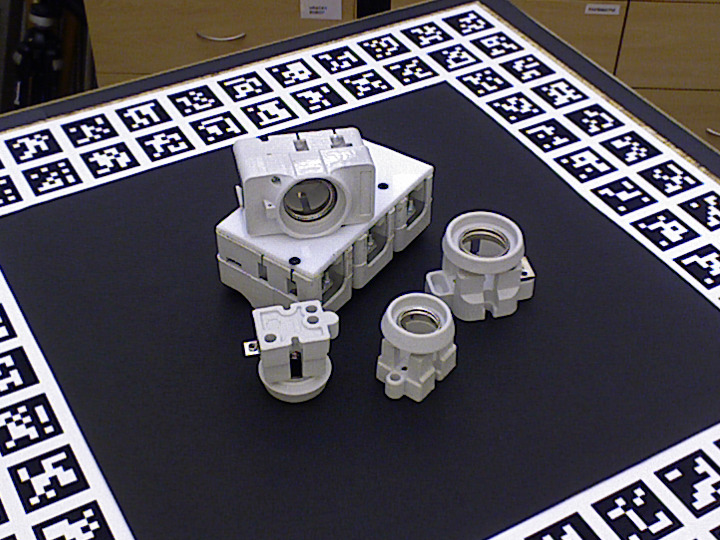}
        \end{subfigure}
    \begin{subfigure}[b]{0.23\textwidth}
        \includegraphics[width=\columnwidth]{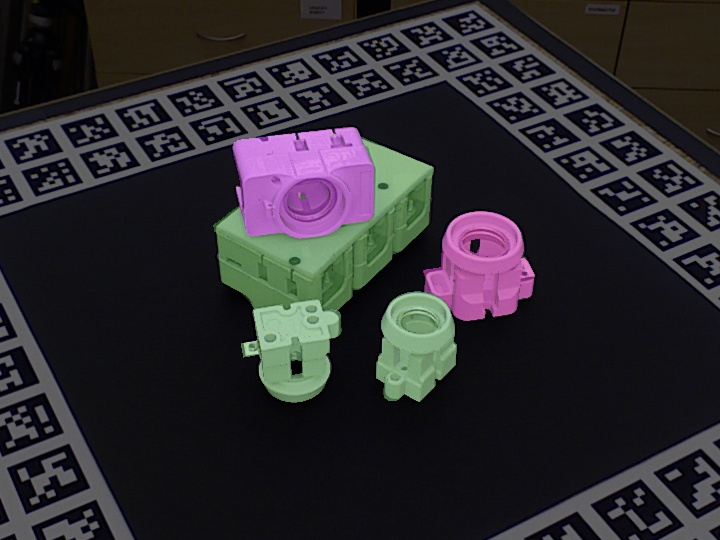}
        \end{subfigure}
    \begin{subfigure}[b]{0.23\textwidth}
        \includegraphics[width=\columnwidth]{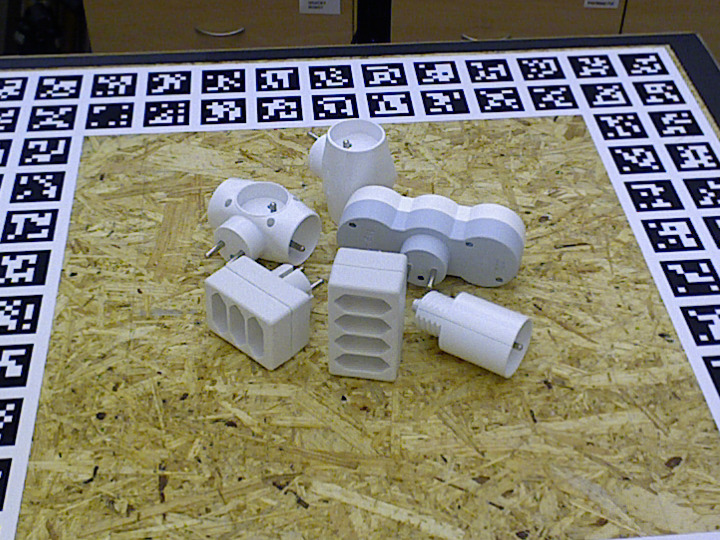}
        \end{subfigure}
    \begin{subfigure}[b]{0.23\textwidth}
        \includegraphics[width=\columnwidth]{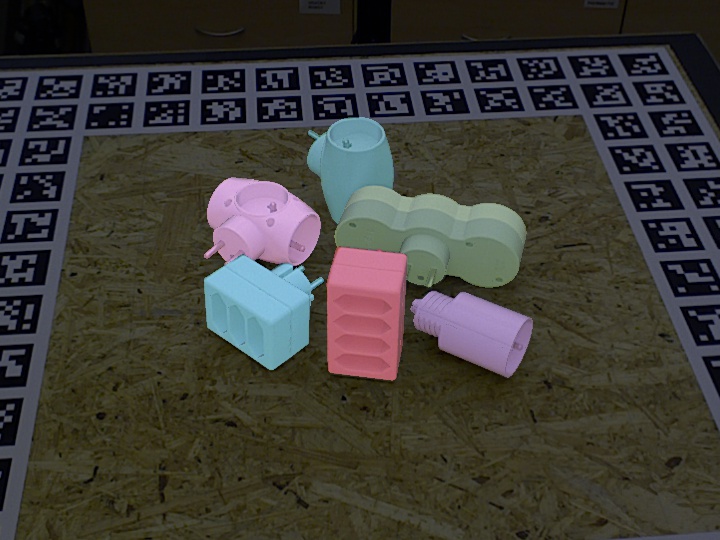}
        \end{subfigure}
    \begin{subfigure}[b]{0.23\textwidth}
        \includegraphics[width=\columnwidth]{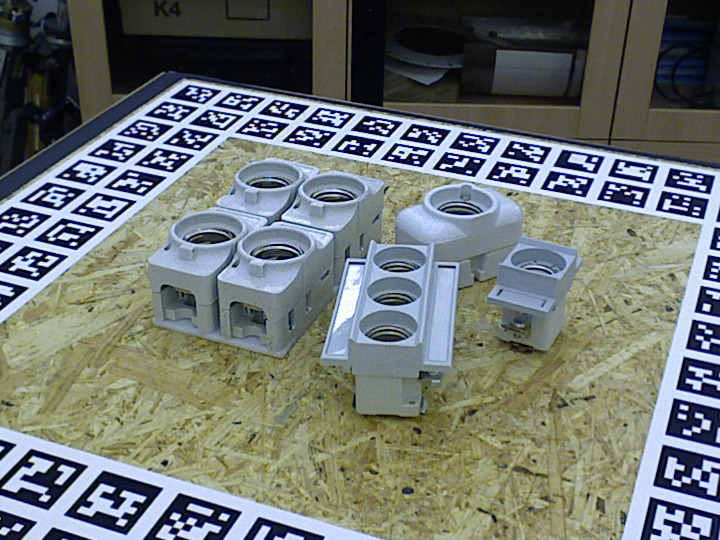}
        \end{subfigure}
    \begin{subfigure}[b]{0.23\textwidth}
        \includegraphics[width=\columnwidth]{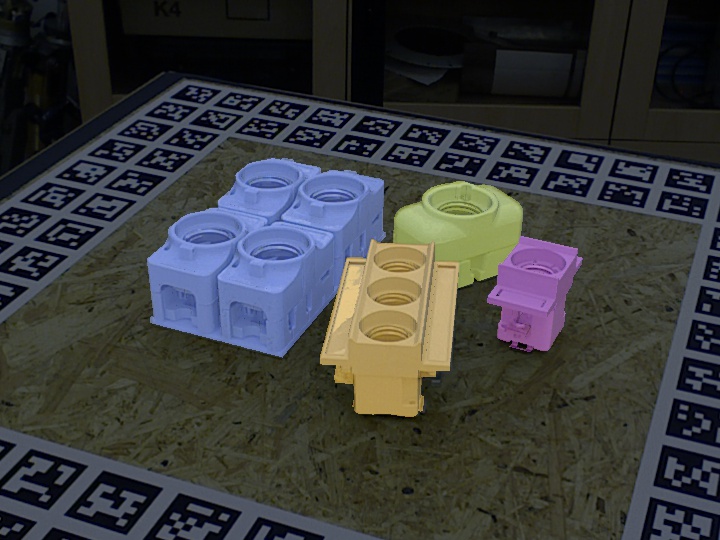}
        \end{subfigure}
    \begin{subfigure}[b]{0.23\textwidth}
        \includegraphics[width=\columnwidth]{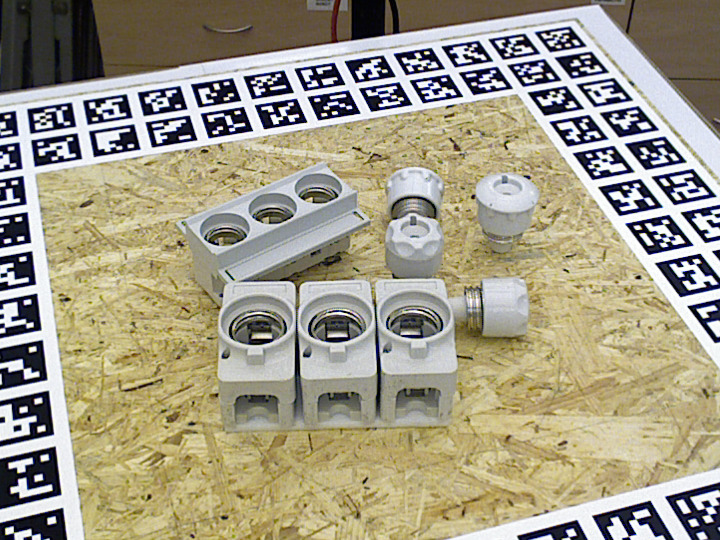}
        \end{subfigure}
    \begin{subfigure}[b]{0.23\textwidth}
        \includegraphics[width=\columnwidth]{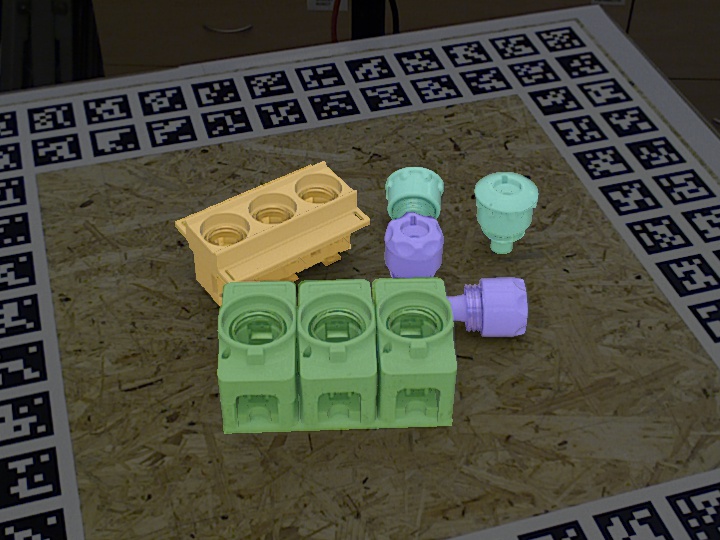}
        \end{subfigure}
    \begin{subfigure}[b]{0.23\textwidth}
        \includegraphics[width=\columnwidth]{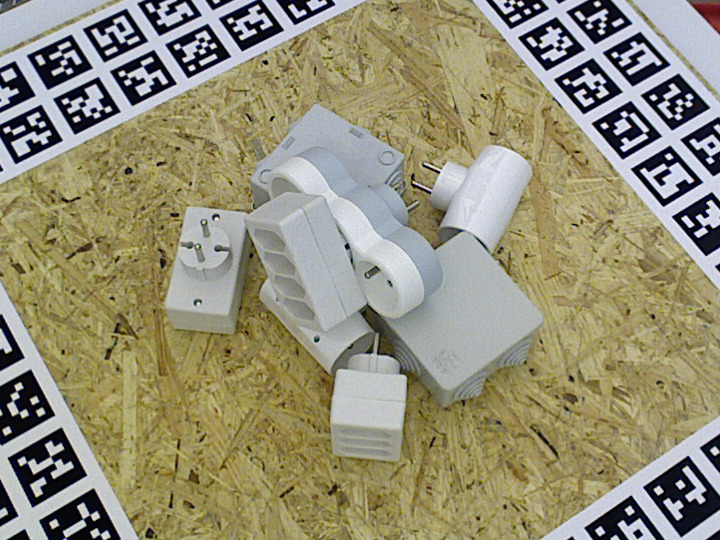}
        \end{subfigure}
    \begin{subfigure}[b]{0.23\textwidth}
        \includegraphics[width=\columnwidth]{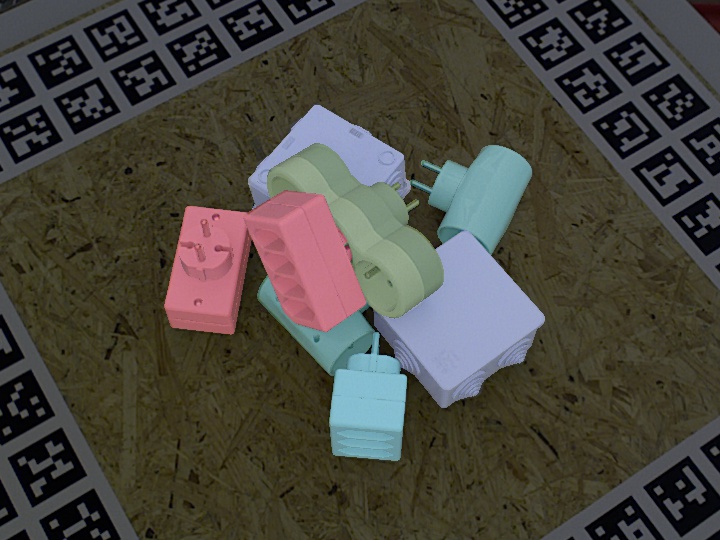}
        \end{subfigure}
    \begin{subfigure}[b]{0.23\textwidth}
        \includegraphics[width=\columnwidth]{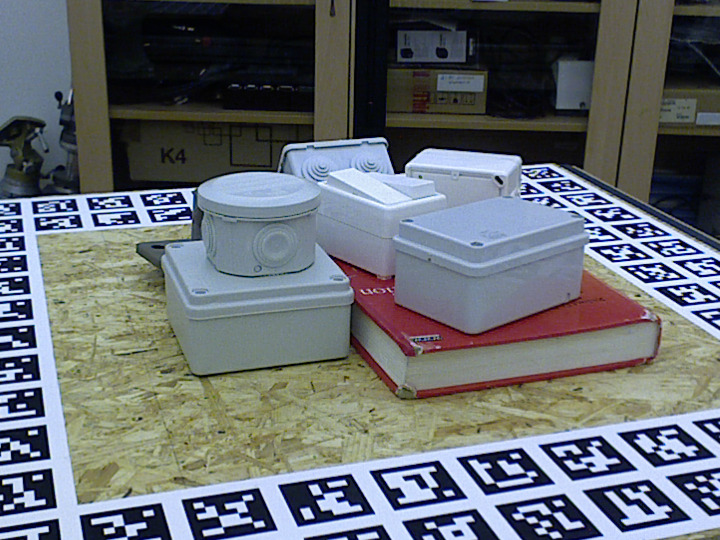}
        \end{subfigure}
    \begin{subfigure}[b]{0.23\textwidth}
        \includegraphics[width=\columnwidth]{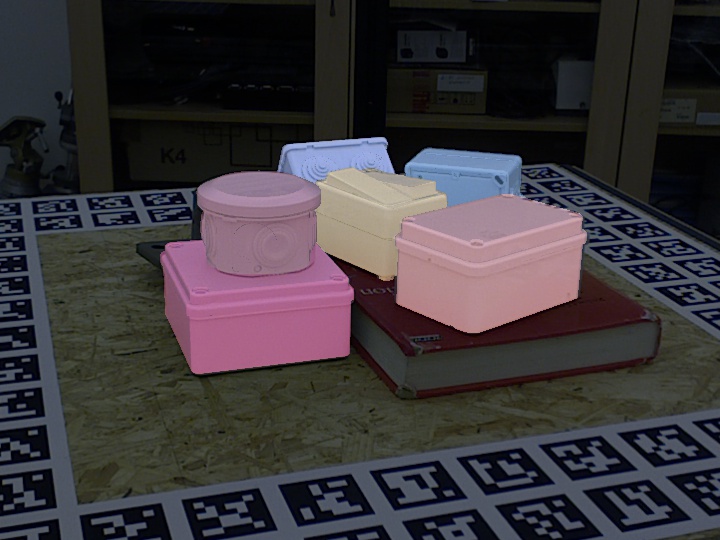}
        \end{subfigure}
    \begin{subfigure}[b]{0.23\textwidth}
        \includegraphics[width=\columnwidth]{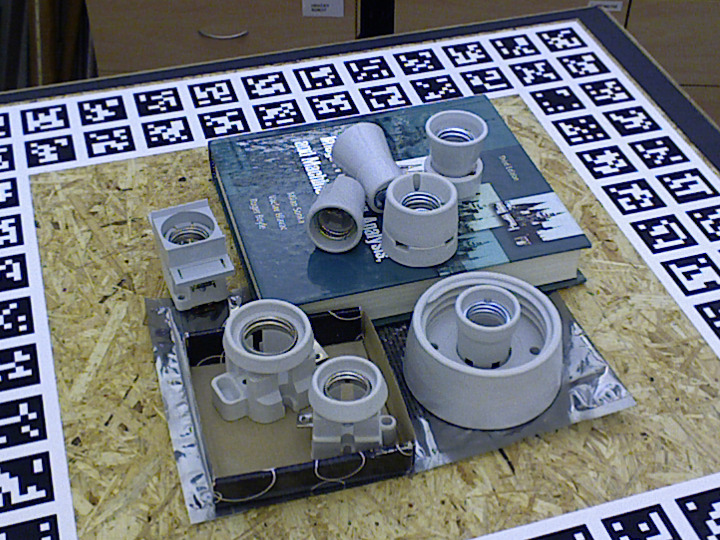}
        \caption{}
        \end{subfigure}
    \begin{subfigure}[b]{0.23\textwidth}
        \includegraphics[width=\columnwidth]{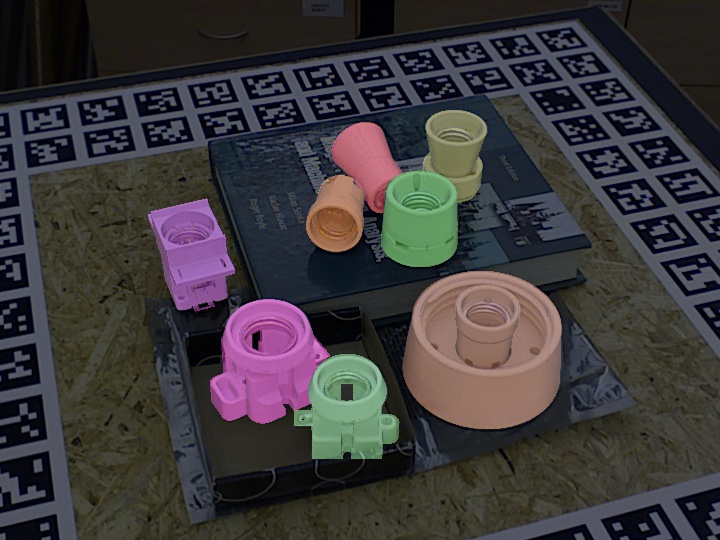}
        \caption{}
        \end{subfigure}
    \begin{subfigure}[b]{0.23\textwidth}
        \includegraphics[width=\columnwidth]{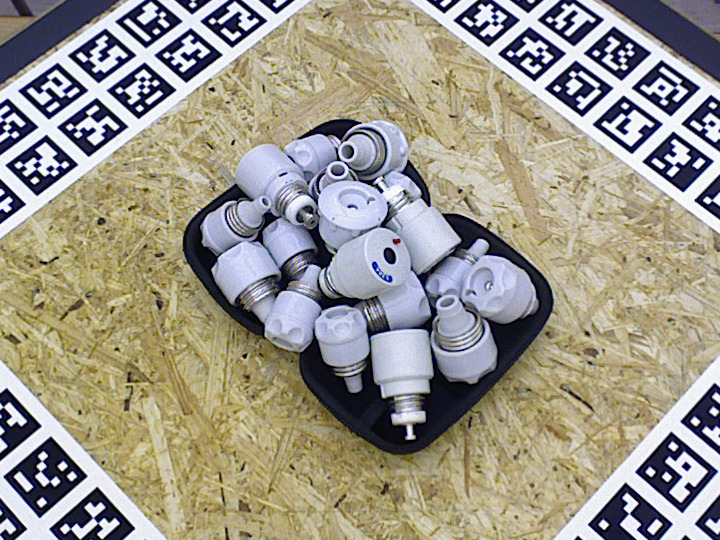}
        \caption{}
        \end{subfigure}
    \begin{subfigure}[b]{0.23\textwidth}
        \includegraphics[width=\columnwidth]{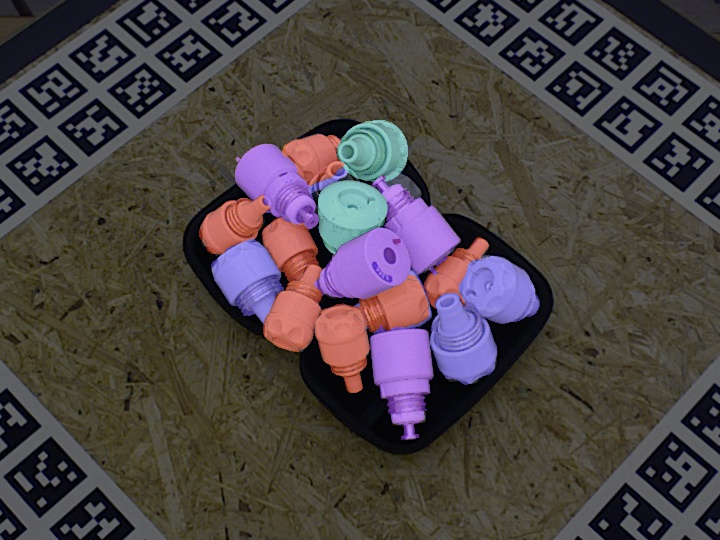}
        \caption{}
        \end{subfigure}
\caption{\textbf{Qualitative results on T-LESS~\cite{hodan2017tless}}: (a, c) The original images and (b, d) MRC-Net object pose predictions. The object’s 3D model is projected with estimated 6D pose and overlaid on original images with distinct colors. Mask RCNN~\cite{he2017mask,labb2020cosypose} detection is used. Best viewed when zoomed in.}
\label{fig:visual_tless}
\end{figure*}

\begin{figure*}[h!]
\centering
    \begin{subfigure}[b]{0.23\textwidth}
        \includegraphics[width=\columnwidth]{figs/vis_est_poses_CVPR_selected_Overlay/ycbv_orig_000048_001075.jpg}
        \end{subfigure}
    \begin{subfigure}[b]{0.23\textwidth}
        \includegraphics[width=\columnwidth]{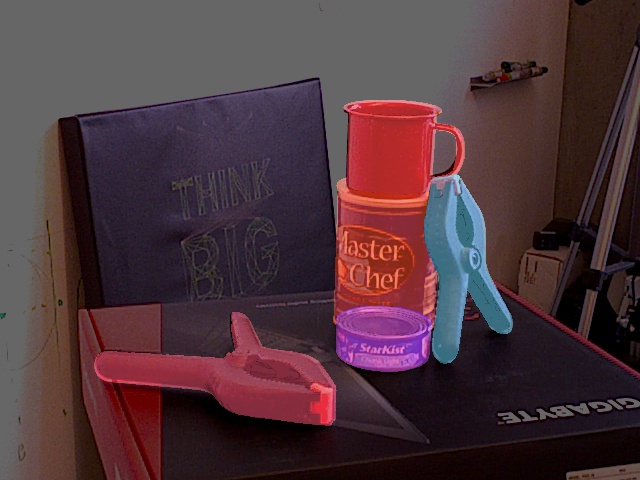}
        \end{subfigure}
    \begin{subfigure}[b]{0.23\textwidth}
        \includegraphics[width=\columnwidth]{figs/vis_est_poses_CVPR_selected_Overlay/ycbv_orig_000049_000558.jpg}
        \end{subfigure}
    \begin{subfigure}[b]{0.23\textwidth}
        \includegraphics[width=\columnwidth]{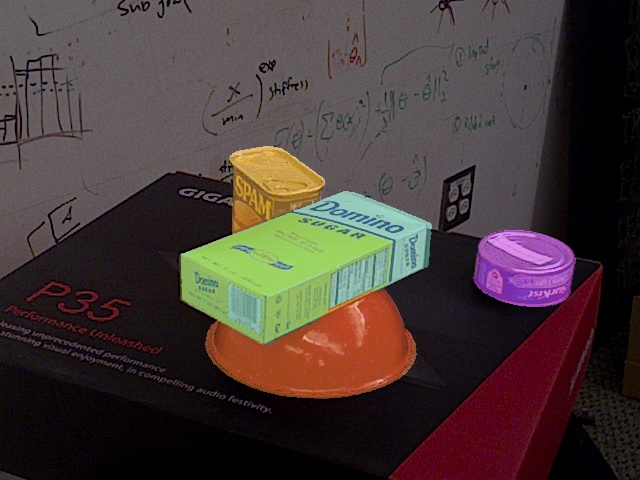}
        \end{subfigure}
    \begin{subfigure}[b]{0.23\textwidth}
        \includegraphics[width=\columnwidth]{figs/vis_est_poses_CVPR_selected_Overlay/ycbv_orig_000050_001044.jpg}
        \end{subfigure}
    \begin{subfigure}[b]{0.23\textwidth}
        \includegraphics[width=\columnwidth]{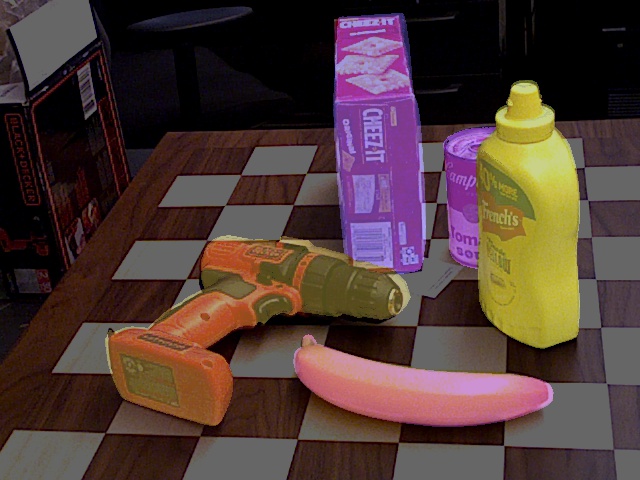}
        \end{subfigure}
    \begin{subfigure}[b]{0.23\textwidth}
        \includegraphics[width=\columnwidth]{figs/vis_est_poses_CVPR_selected_Overlay/ycbv_orig_000051_001584.jpg}
        \end{subfigure}
    \begin{subfigure}[b]{0.23\textwidth}
        \includegraphics[width=\columnwidth]{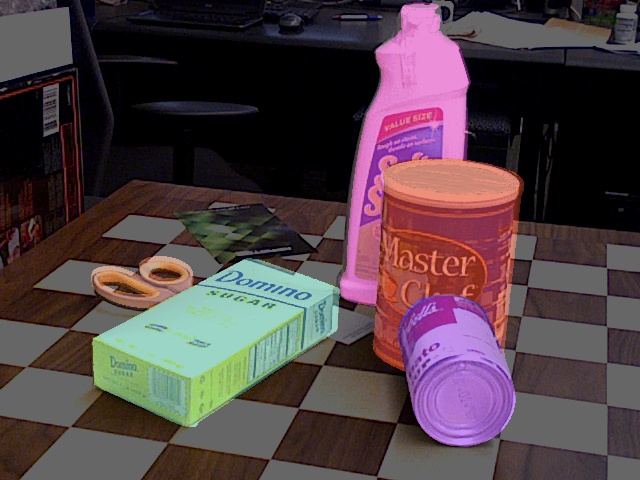}
        \end{subfigure}
    \begin{subfigure}[b]{0.23\textwidth}
        \includegraphics[width=\columnwidth]{figs/vis_est_poses_CVPR_selected_Overlay/ycbv_orig_000052_000175.jpg}
        \end{subfigure}
    \begin{subfigure}[b]{0.23\textwidth}
        \includegraphics[width=\columnwidth]{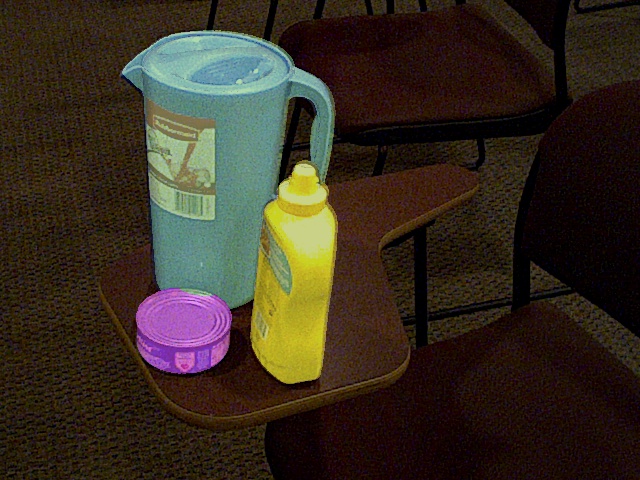}
        \end{subfigure}
    \begin{subfigure}[b]{0.23\textwidth}
        \includegraphics[width=\columnwidth]{figs/vis_est_poses_CVPR_selected_Overlay/ycbv_orig_000053_000692.jpg}
        \end{subfigure}
    \begin{subfigure}[b]{0.23\textwidth}
        \includegraphics[width=\columnwidth]{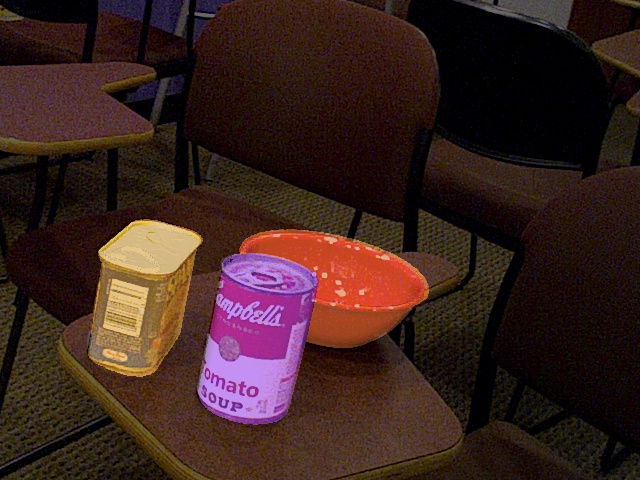}
        \end{subfigure}
    \begin{subfigure}[b]{0.23\textwidth}
        \includegraphics[width=\columnwidth]{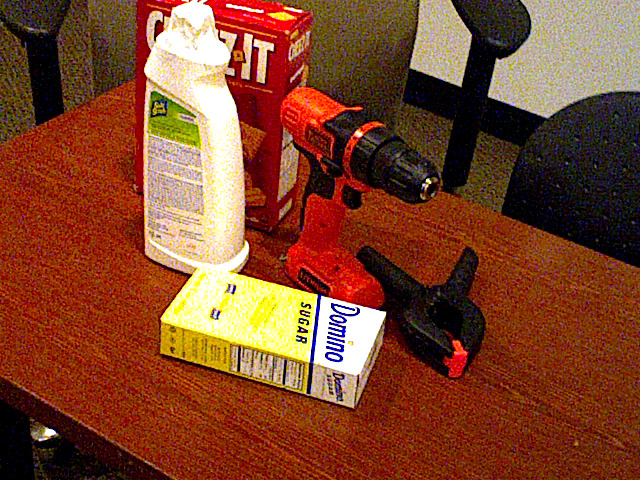}
        \end{subfigure}
    \begin{subfigure}[b]{0.23\textwidth}
        \includegraphics[width=\columnwidth]{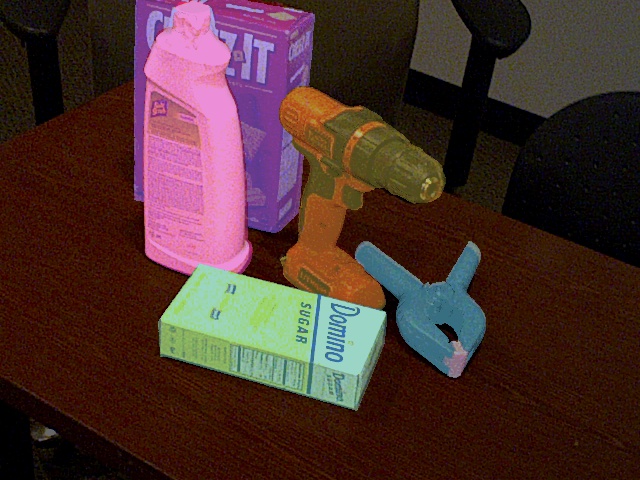}
        \end{subfigure}
    \begin{subfigure}[b]{0.23\textwidth}
        \includegraphics[width=\columnwidth]{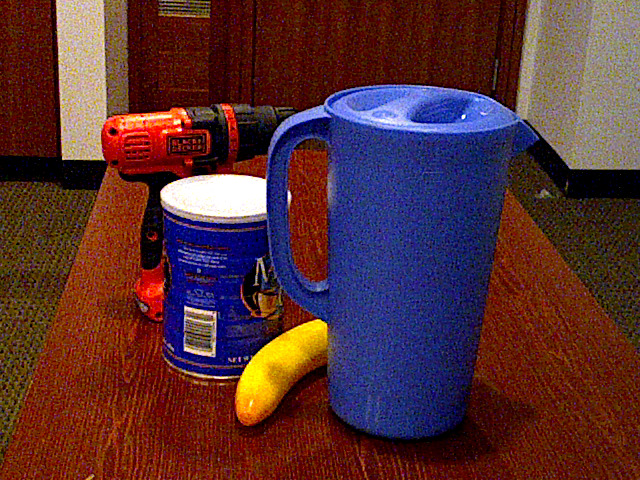}
        \end{subfigure}
    \begin{subfigure}[b]{0.23\textwidth}
        \includegraphics[width=\columnwidth]{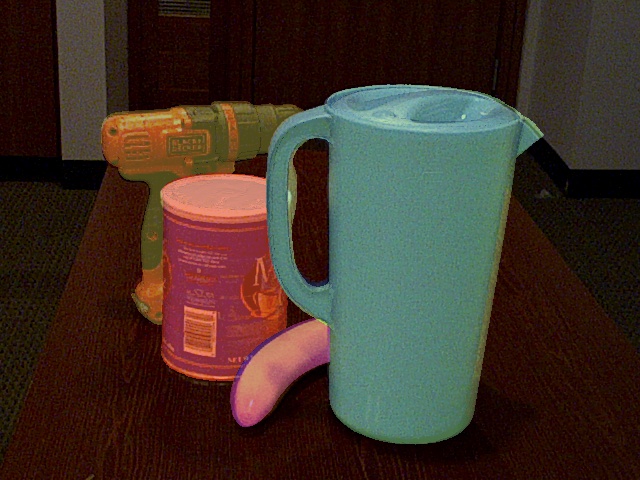}
        \end{subfigure}
    \begin{subfigure}[b]{0.23\textwidth}
        \includegraphics[width=\columnwidth]{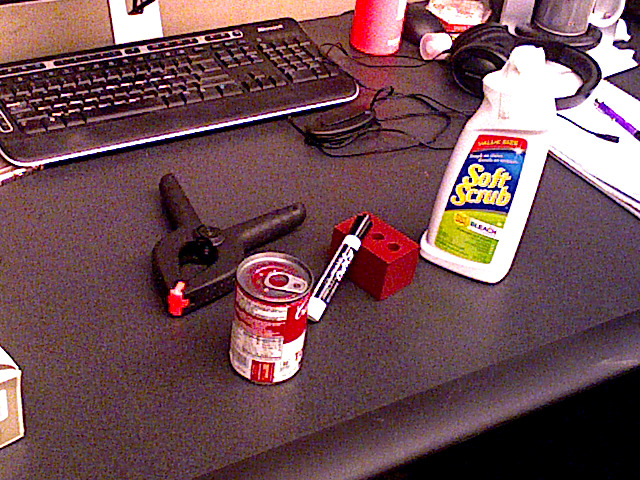}
        \caption{}
        \end{subfigure}
    \begin{subfigure}[b]{0.23\textwidth}
        \includegraphics[width=\columnwidth]{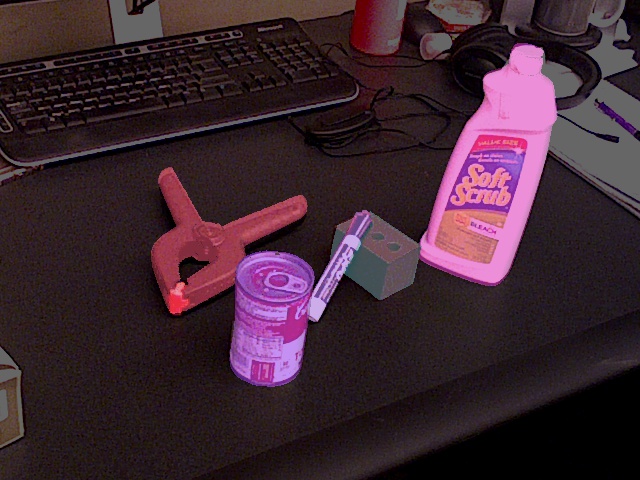}
        \caption{}
        \end{subfigure}
    \begin{subfigure}[b]{0.23\textwidth}
        \includegraphics[width=\columnwidth]{figs/vis_est_poses_CVPR_selected_Overlay/ycbv_orig_000059_001039.jpg}
        \caption{}
        \end{subfigure}
    \begin{subfigure}[b]{0.23\textwidth}
        \includegraphics[width=\columnwidth]{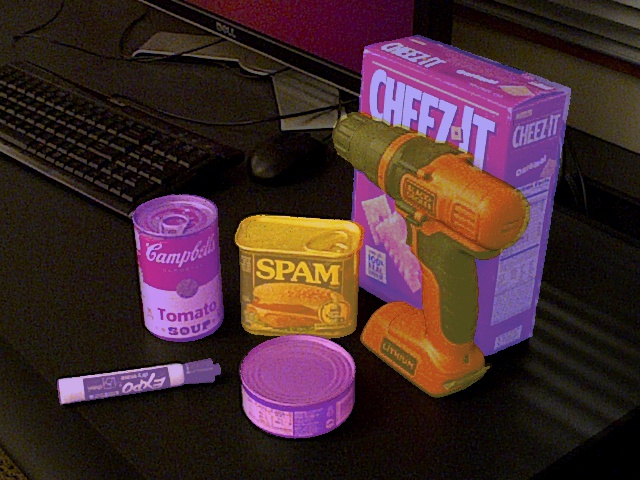}
        \caption{}
        \end{subfigure}
\caption{\textbf{Qualitative results on YCB-V~\cite{xiang2017posecnn}}: (a, c) The original images and (b, d) MRC-Net object pose predictions. Mask RCNN~\cite{he2017mask,labb2020cosypose} detection is used. Best viewed when zoomed in.}
\label{fig:visual_ycbv}
\end{figure*}

\begin{figure*}[h!]
\centering
    \begin{subfigure}[b]{0.23\textwidth}
        \includegraphics[width=\columnwidth]{figs/vis_est_poses_CVPR_selected_Overlay/lmo_orig_000002_000008.jpg}
        \end{subfigure}
    \begin{subfigure}[b]{0.23\textwidth}
        \includegraphics[width=\columnwidth]{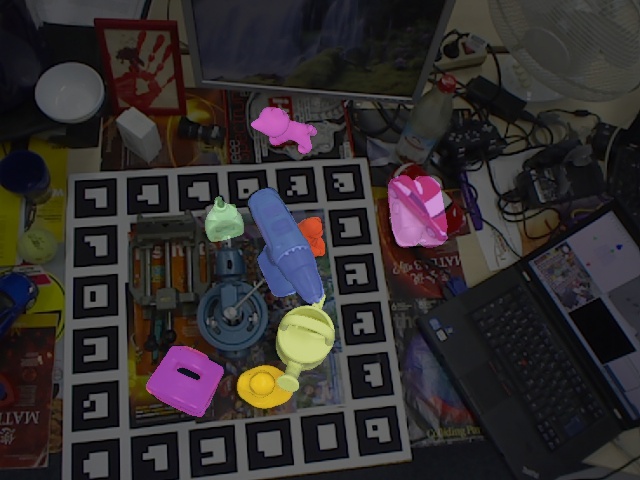}
        \end{subfigure}
    \begin{subfigure}[b]{0.23\textwidth}
        \includegraphics[width=\columnwidth]{figs/vis_est_poses_CVPR_selected_Overlay/lmo_orig_000002_000039.jpg}
        \end{subfigure}
    \begin{subfigure}[b]{0.23\textwidth}
        \includegraphics[width=\columnwidth]{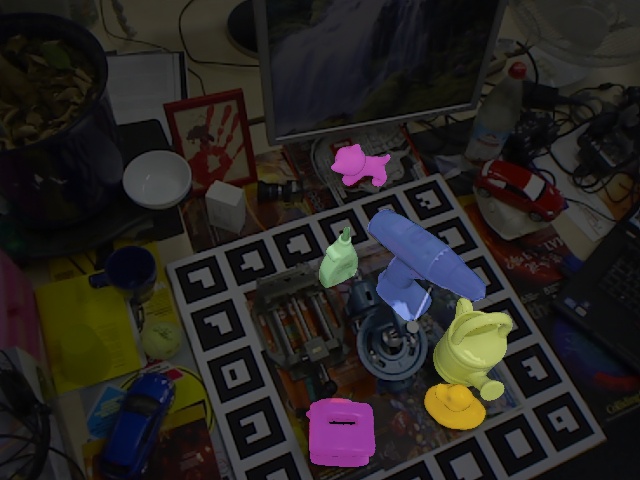}
        \end{subfigure}
    \begin{subfigure}[b]{0.23\textwidth}
        \includegraphics[width=\columnwidth]{figs/vis_est_poses_CVPR_selected_Overlay/lmo_orig_000002_000124.jpg}
        \end{subfigure}
    \begin{subfigure}[b]{0.23\textwidth}
        \includegraphics[width=\columnwidth]{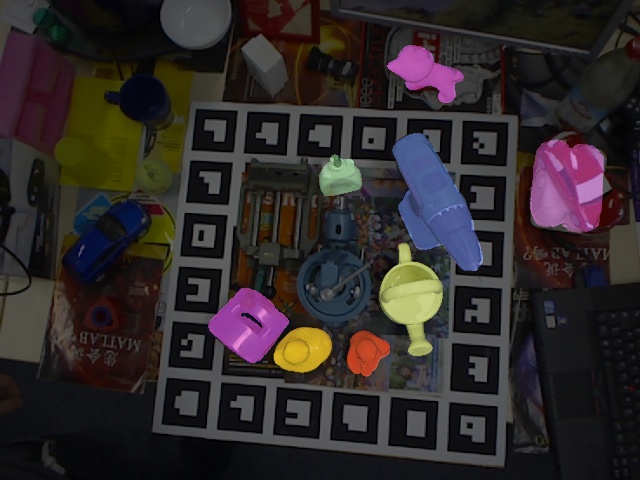}
        \end{subfigure}
    \begin{subfigure}[b]{0.23\textwidth}
        \includegraphics[width=\columnwidth]{figs/vis_est_poses_CVPR_selected_Overlay/lmo_orig_000002_000243.jpg}
        \end{subfigure}
    \begin{subfigure}[b]{0.23\textwidth}
        \includegraphics[width=\columnwidth]{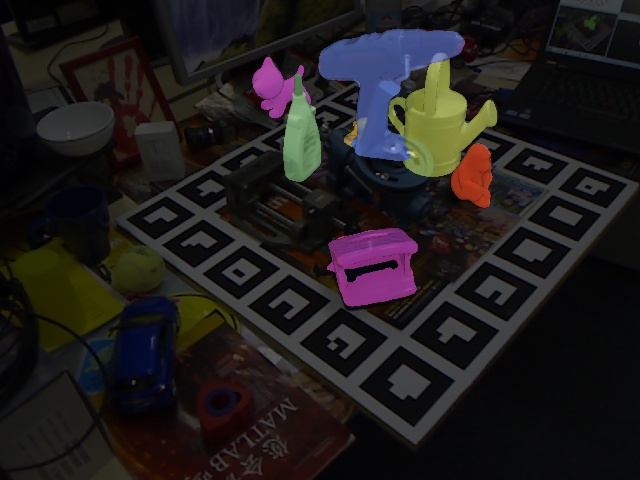}
        \end{subfigure}
    \begin{subfigure}[b]{0.23\textwidth}
        \includegraphics[width=\columnwidth]{figs/vis_est_poses_CVPR_selected_Overlay/lmo_orig_000002_000307.jpg}
        \end{subfigure}
    \begin{subfigure}[b]{0.23\textwidth}
        \includegraphics[width=\columnwidth]{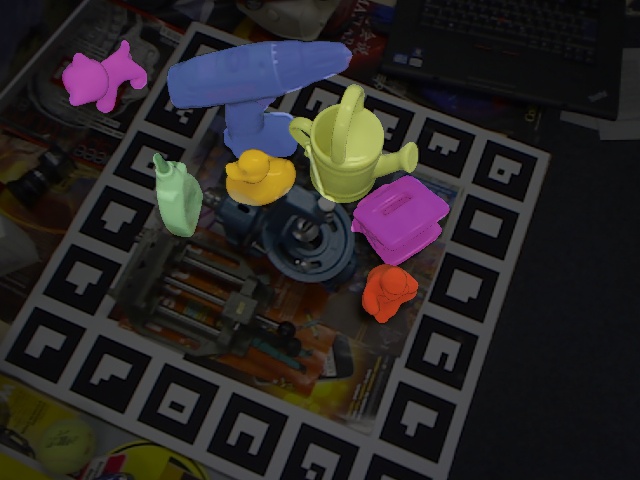}
        \end{subfigure}
    \begin{subfigure}[b]{0.23\textwidth}
        \includegraphics[width=\columnwidth]{figs/vis_est_poses_CVPR_selected_Overlay/lmo_orig_000002_000362.jpg}
        \end{subfigure}
    \begin{subfigure}[b]{0.23\textwidth}
        \includegraphics[width=\columnwidth]{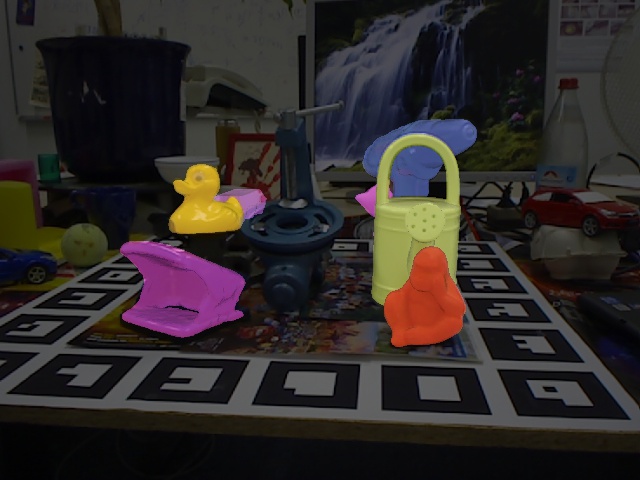}
        \end{subfigure}
    \begin{subfigure}[b]{0.23\textwidth}
        \includegraphics[width=\columnwidth]{figs/vis_est_poses_CVPR_selected_Overlay/lmo_orig_000002_000473.jpg}
        \end{subfigure}
    \begin{subfigure}[b]{0.23\textwidth}
        \includegraphics[width=\columnwidth]{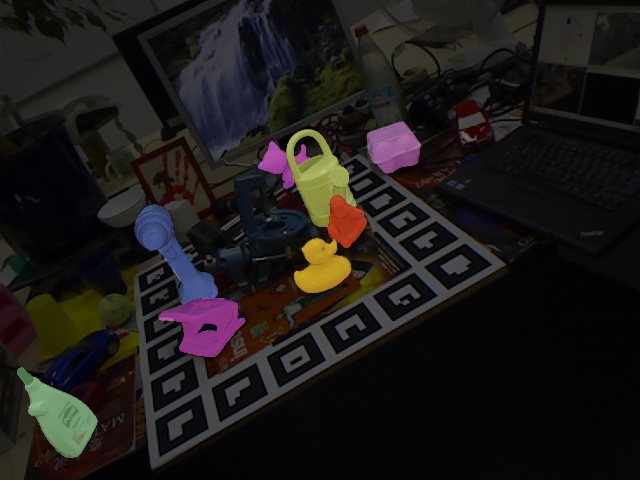}
        \end{subfigure}
    \begin{subfigure}[b]{0.23\textwidth}
        \includegraphics[width=\columnwidth]{figs/vis_est_poses_CVPR_selected_Overlay/lmo_orig_000002_000669.jpg}
        \end{subfigure}
    \begin{subfigure}[b]{0.23\textwidth}
        \includegraphics[width=\columnwidth]{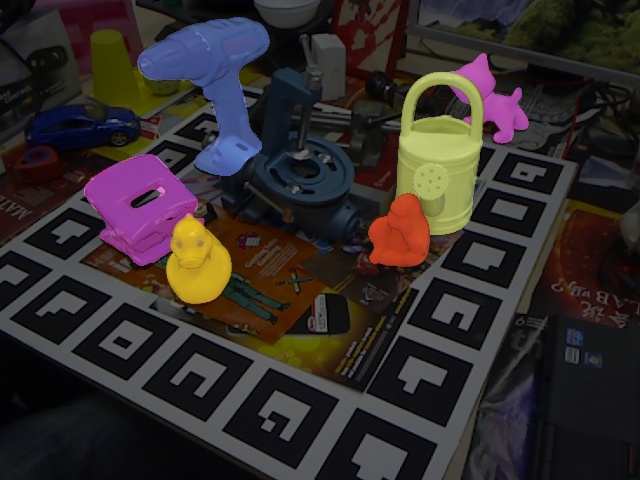}
        \end{subfigure}
    \begin{subfigure}[b]{0.23\textwidth}
        \includegraphics[width=\columnwidth]{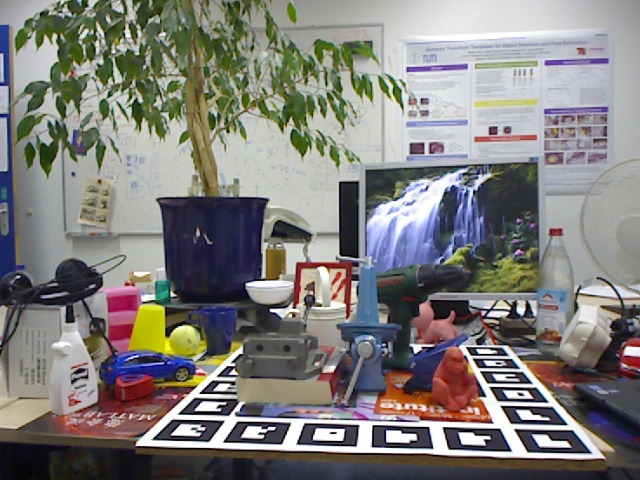}
        \caption{}
        \end{subfigure}
    \begin{subfigure}[b]{0.23\textwidth}
        \includegraphics[width=\columnwidth]{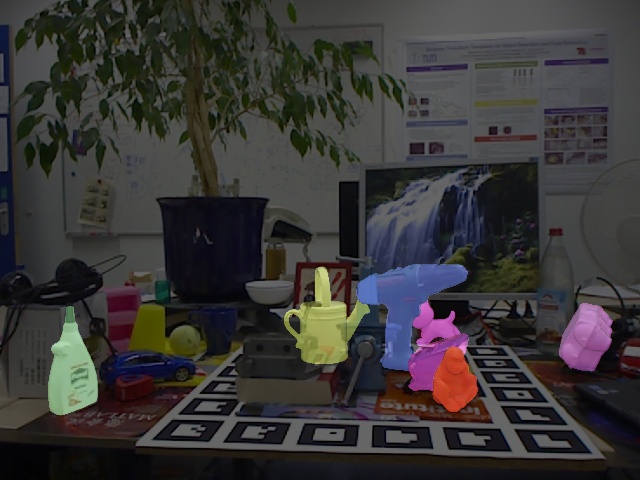}
        \caption{}
        \end{subfigure}
    \begin{subfigure}[b]{0.23\textwidth}
        \includegraphics[width=\columnwidth]{figs/vis_est_poses_CVPR_selected_Overlay/lmo_orig_000002_000850.jpg}
        \caption{}
        \end{subfigure}
    \begin{subfigure}[b]{0.23\textwidth}
        \includegraphics[width=\columnwidth]{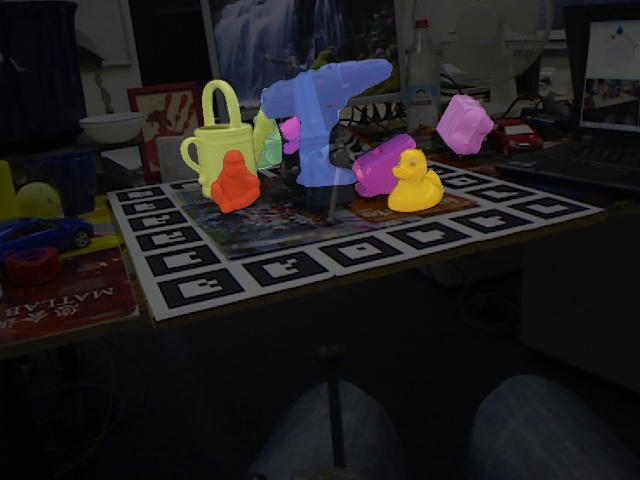}
        \caption{}
        \end{subfigure}
\caption{\textbf{Qualitative results on LM-O~\cite{brachmann2014learning}}: (a, c) The original images and (b, d) MRC-Net object pose predictions. Mask RCNN detection~\cite{he2017mask,labb2020cosypose} is used. Best viewed when zoomed in.}
\label{fig:visual_lmo}
\end{figure*}

Visual examples from the T-LESS, YCB-V and LM-O datasets are presented in Fig.~\ref{fig:visual_tless}, Fig.~\ref{fig:visual_ycbv} and Fig.~\ref{fig:visual_lmo}, respectively. These examples demonstrate MRC-Net's ability to accurately predicting object poses in challenging scenarios such as heavy occlusions, diverse viewpoints, and distracting background clutter. 

\begin{figure*}[h]
\centering
\begin{subfigure}[b]{0.43\textwidth}
        \includegraphics[width=\columnwidth]{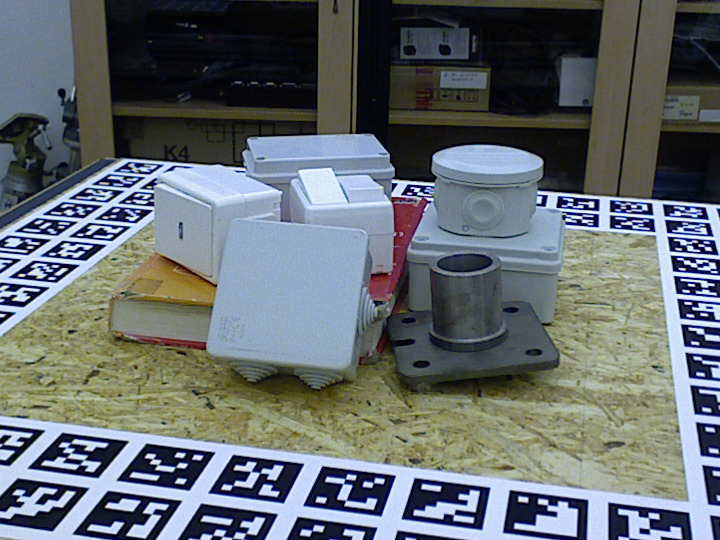}
        \caption{}
        \end{subfigure}
    \begin{subfigure}[b]{0.43\textwidth}
        \includegraphics[width=\columnwidth]{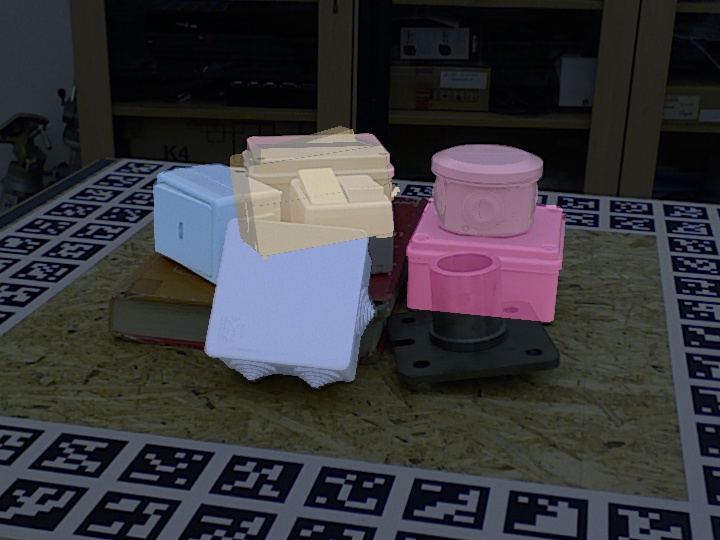}
        \caption{}
        \end{subfigure}
    \begin{subfigure}[b]{0.43\textwidth}
        \includegraphics[width=\columnwidth]
        {figs/vis_est_poses_CVPR_selected_Overlay_failures/ycbv_orig_000049_000091.jpg}
        \caption{}
        \end{subfigure}
    \begin{subfigure}[b]{0.43\textwidth}
        \includegraphics[width=\columnwidth]
        {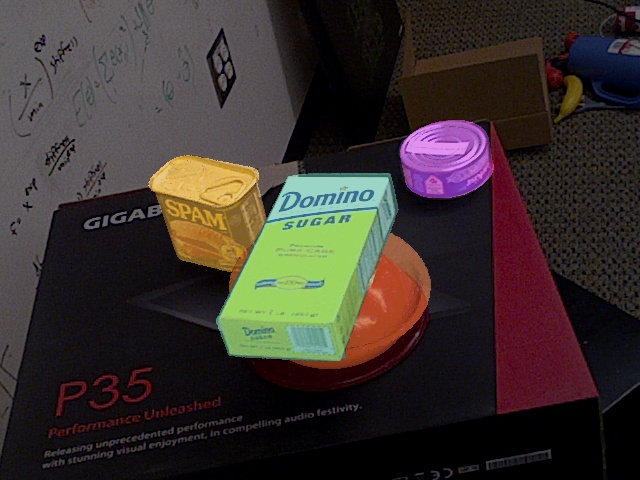}
        \caption{}
        \end{subfigure}
            \begin{subfigure}[b]{0.43\textwidth}
        \includegraphics[width=\columnwidth]
        {figs/vis_est_poses_CVPR_selected_Overlay_failures/lmo_orig_000002_000877.jpg}
        \caption{}
        \end{subfigure}
    \begin{subfigure}[b]{0.43\textwidth}
        \includegraphics[width=\columnwidth]
        {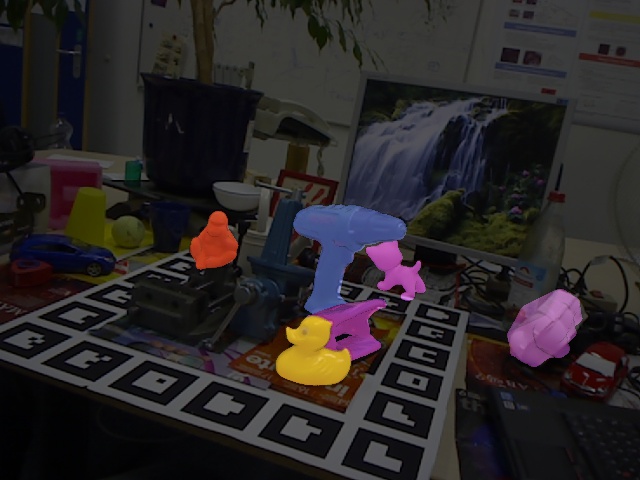}
        \caption{}
        \end{subfigure}
\caption{\textbf{Failure Examples}: (a), (c), and (e) show the original images. (b), (d), and (f) represent MRC-net predictions. Observe the flipped object pose induced by heavy occlusion in the center in (b), the upside-down red bowl in (d), and the inaccurately rotated eggbox in (f).}
\label{fig:failures}
\end{figure*}

We present typical failure cases per dataset in Fig.~\ref{fig:failures}. In Fig.~\ref{fig:failures} (b), the object in the center has an inaccurately estimated pose due to heavy occlusion. In Fig.~\ref{fig:failures} (d), the object 0024\_bowl in the training set predominantly faces upward, and occasional incorrect flips of the bowl occur due to such pose imbalance. Fig.~\ref{fig:failures} (f) illustrates the eggbox object with incorrect rotations, partly attributed to inaccuracies in the CAD model.

\begin{table*}[h!]
\adjustbox{max width=\textwidth}{
    \begin{tabular}{l|cc|cc|cc|cc|cc|cc}
      \toprule
Method              & \multicolumn{2}{c}{PoseCNN~\cite{xiang2017posecnn}} & \multicolumn{2}{c}{GDR-Net~\cite{wang2021gdr}} & \multicolumn{2}{c}{ZebraPose~\cite{su2022zebrapose}} & \multicolumn{2}{c}{CheckerPose~\cite{lian2023checkerpose}} & \multicolumn{2}{c}{MRC-Net} & \multicolumn{2}{c}{MRC-Net}\\
\midrule
Detection & \multicolumn{2}{c}{Built-in} & \multicolumn{2}{c}{Faster-RCNN~\cite{ren2015faster,li2019cdpn}} & \multicolumn{2}{c}{FCOS~\cite{tian2019fcos,li2019cdpn}} & \multicolumn{2}{c}{FCOS~\cite{tian2019fcos,li2019cdpn}} & \multicolumn{2}{c}{Mask RCNN~\cite{he2017mask,labb2020cosypose}} & \multicolumn{2}{c}{FCOS~\cite{tian2019fcos,li2019cdpn}}\\
\midrule
Metric              & \makecell{AUC of\\ADD-S} & \makecell{AUC of\\ADD(-S)} & \makecell{AUC of\\ADD-S} & \makecell{AUC of\\ADD(-S)} & \makecell{AUC of\\ADD-S} & \makecell{AUC of\\ADD(-S)} & \makecell{AUC of\\ADD-S} & \makecell{AUC of\\ADD(-S)} & \makecell{AUC of\\ADD-S} & \makecell{AUC of\\ADD(-S)} & \makecell{AUC of\\ADD-S} & \makecell{AUC of\\ADD(-S)}\\
\midrule
002\_master\_chef\_can & 84.0 & 50.9 & 96.3 & 65.2 & 93.7 & 75.4 & 87.5 & 67.7 & \tcr{98.2} & \tcb{98.2} & 98.0 & 98.0 \\
      003\_cracker\_box      & 76.9 & 51.7 & 97.0 & 88.8 & 93.0 & 87.8 & 93.2 & 86.7 & \tcr{100.0} & 98.6 & 99.9 &  \tcb{98.8}\\
      004\_sugar\_box        & 84.3 & 68.6 & 98.9 & 95.0 & 95.1 & 90.9 & 95.9 & 91.7 & \tcr{100.0} & 98.9 & \tcr{100.0} & \tcb{99.0} \\
      005\_tomato\_soup\_can  & 80.9 & 66.0 & \tcr{96.5} & 91.9 & 94.4 & 90.1 & 94.0 & 89.9 & 95.1 & 92.6 & 96.0 & \tcb{93.5} \\
      006\_mustard\_bottle   & 90.2 & 79.9 & \tcr{100.0}  & 92.8 & 96.0 & 92.6 & 95.7 & 90.9 & 99.9 & \tcb{98.9} & 99.7 & 98.1\\
      007\_tuna\_fish\_can    & 87.9 & 70.4 & \tcr{99.4} & 94.2 & 96.9 & 92.6 & 97.5 & \tcb{94.4} & 97.2 & 87.3 & 97.3 & 87.5 \\
      008\_pudding\_box      & 79.0 & 62.9 & 64.6 & 44.7 & 97.2 & 95.3 & 94.9 & 91.5 & \tcr{99.6} & \tcb{98.4} & 99.2 & 97.9 \\
      009\_gelatin\_box      & 87.1 & 75.2 & 97.1 & 92.5 & 96.8 & 94.8 & 96.1 & 93.4 & \tcr{98.9} & \tcb{96.7} & \tcr{98.9} & 96.3 \\
      010\_potted\_meat\_can  & 78.5 & 59.6 & 86.0 & 80.2 & \tcr{91.7} & \tcb{83.6} & 86.4 & 80.4 & 79.2 & 75.2 & 84.1 & 79.0 \\
      011\_banana           & 85.9 & 72.3 & 96.3 & 85.8 & 92.6 & 84.6 & 95.7 & 90.1 & \tcr{98.9} & \tcb{93.0} & 98.3 & 91.9\\
      019\_pitcher\_base     & 76.8 & 52.5 & 99.9 & 98.5 & 96.4 & 93.4 & 95.8 & 91.9 & 99.9 & \tcb{99.6} & \tcr{100.0} & 99.3\\
      021\_bleach\_cleanser  & 71.9 & 50.5 & \tcr{94.2} & 84.3 & 89.5 & 80.0 & 90.6 & 83.2 & 92.5 & 83.2 & 93.7 & \tcb{85.2} \\
      024\_bowl             & 69.7 & 69.7 & 85.7 & 85.7 & 37.1 & 37.1 & 82.5 & 82.5 & \tcr{99.0} & \tcb{99.0} & 98.7 & 98.7 \\
      025\_mug              & 78.0 & 57.7 & 99.6 & 94.0 & 96.1 & 90.8 & 96.9 & 92.7 & \tcr{99.7} & \tcb{99.7} & 99.4 & 99.4 \\
      035\_power\_drill      & 72.8 & 55.1 & 97.5 & 90.1 & 95.0 & 89.7 & 94.7 & 88.8 & \tcr{99.9} & \tcb{98.1} & 99.8 & 97.9\\
      036\_wood\_block       & 65.8 & 65.8 & 82.5 & 82.5 & \tcr{84.5} & \tcb{84.5} & 68.3 & 68.3 & 83.9 & 83.9 & 84.4 & 84.4\\
      037\_scissors         & 56.2 & 35.8 & 63.8 & 49.5 & 92.5 & 84.5 & 91.7 & 81.6 & 90.4 & 78.0 & \tcr{94.5} & \tcb{85.9} \\
      040\_large\_marker     & 71.4 & 58.0 & 88.0 & 76.1 & 80.4 & 69.5 & 83.3 & 72.3 & 96.4 & 96.4 & \tcr{97.3} & \tcb{97.3} \\
      051\_large\_clamp      & 49.9 & 49.9 & 89.3 & 89.3 & 85.6 & 85.6 & 90.0 & 90.0 & 93.2 & 93.2 & \tcr{97.1} & \tcb{97.1}\\
      052\_extra\_large\_clamp & 47.0 & 47.0 & 93.5 & 93.5 & 92.5 & 92.5 & 91.6 & 91.6 & 62.1 & 62.1 & \tcr{98.3} & \tcb{98.3}\\
      061\_foam\_brick       & 87.8 & 87.8 & \tcr{96.9} & \tcb{96.9} & 95.3 & 95.3 & 94.1 & 94.1 & 95.5 & 95.5 & 96.7 & 96.7 \\
\midrule
Mean                & 75.9 & 61.3 & 91.6 & 84.3 & 90.1 & 85.3 & 91.3 & 86.4 & 94.3     & 91.7 & \tcr{96.7} & \tcb{94.3} \\
\bottomrule
\end{tabular}
}
\caption{Detailed results on YCB-V~\cite{xiang2017posecnn} w.r.t. AUC of ADD-S and AUC of ADD(-S). We highlight the best AUC of ADD-S results in \tcr{red}, and the best AUC of ADD(-S) in \tcb{blue}.}\label{tab:ycbv_per_obj}
\end{table*}

\clearpage
\begin{table*}
    \centering
    \adjustbox{max width=\textwidth}{
    \begin{tabular}{l|cccccccc}
        \toprule
        Method & SegDriven~\cite{hu2019segmentation} & GDR~\cite{wang2021gdr} & Zebra~\cite{su2022zebrapose} & CheckerPose~\cite{lian2023checkerpose} & MRC-Net & MRC-Net \\
        \midrule
        Detection & Built-in & Faster RCNN~\cite{ren2015faster,li2019cdpn} & FCOS~\cite{tian2019fcos,li2019cdpn} & FCOS~\cite{tian2019fcos,li2019cdpn} & Mask RCNN~\cite{he2017mask,labb2020cosypose} & FCOS~\cite{tian2019fcos,li2019cdpn} \\
        \midrule
        002\_master\_chef\_can & 33.0 & 41.5 & 62.6 & 45.9 & \textbf{96.3} & 94.3 \\
        003\_cracker\_box & 44.6 & 83.2 & 98.5 & 94.2 & \textbf{100.0} & \textbf{100.0} \\
        004\_sugar\_box & 75.6 & 91.5 & 96.3 & 98.3 & \textbf{100.0} & \textbf{100.0} \\
        005\_tomato\_soup\_can & 40.8 & 65.9 & 80.5 & 83.2 & 79.9 & \textbf{81.0} \\
        006\_mustard\_bottle & 70.6 & 90.2 & \textbf{100.0} & 99.2 & 98.0 & 96.7\\
        007\_tuna\_fish\_can & 18.1 & 44.2 & 70.5 & \textbf{88.9} & 14.3 & 17.0 \\
        008\_pudding\_box & 12.2 & 2.8 & \textbf{99.5} & 86.5 & 92.0 & 93.3 \\
        009\_gelatin\_box & 59.4 & 61.7 & \textbf{97.2} & 86.0 & 69.3 & 70.7\\
        010\_potted\_meat\_can & 33.3 & 64.9 & \textbf{76.9} & 70.0 & 61.3 & 62.2 \\
        011\_banana & 16.6 & 64.1 & 71.2 & \textbf{96.0} & 86.0 & 82.7 \\
        019\_pitcher\_base & 90.0 & 99.0 & \textbf{100.0} & \textbf{100.0} & 99.6 & \textbf{100.0} \\
        021\_bleach\_cleanser & 70.9 & 73.8 & 75.9 & \textbf{89.8} & 79.7 & 80.7 \\
        024\_bowl & 30.5 & 37.7 & 18.5 & 68.0 & \textbf{92.7} & 90.0 \\
        025\_mug & 40.7 & 61.5 & 77.5 & 89.0 & \textbf{100.0} & 98.0 \\
        035\_power\_drill & 63.5 & 78.5 & 97.4 & 95.9 & \textbf{99.3} & \textbf{99.3} \\
        036\_wood\_block & 27.7 & 59.5 & \textbf{87.6} & 58.7 & 54.7 & 58.7\\
        037\_scissors & 17.1 & 3.9 & \textbf{71.8} & 62.4 & 33.3 & 70.7\\
        040\_large\_marker & 4.8 & 7.4 & 23.3 & 18.8 & 82.7 & \textbf{87.3} \\
        051\_large\_clamp & 25.6 & 69.8 & 87.6 & \textbf{95.4} & 90.7 & 94.7\\
        052\_extra\_large\_clamp & 8.8 & 90.0 & 98.0 & 95.6 & 62.0 & \textbf{99.3}\\
        061\_foam\_brick & 34.7 & 71.9 & \textbf{99.3} & 87.2 & 72.0 & 70.7\\
        \midrule
        Mean & 39.0 & 60.1 & 80.5 & 81.4 & 79.2 & \textbf{83.2}\\
        \bottomrule
    \end{tabular}
    }
    \caption{Detailed results on YCB-V~\cite{xiang2017posecnn} w.r.t. ADD(-S). We highlight the best results in \textbf{bold}.}\label{tab:ycbv_adds}
\end{table*}
\clearpage

\newpage
\afterpage{
{\small
\bibliographystyle{ieeenat_fullname}
\bibliography{egbib}
}
}